\documentclass[letterpaper]{article}
\usepackage{aaai}
\usepackage{times}
\usepackage{helvet}
\usepackage{courier}
\usepackage{enumitem} 
\usepackage{amsmath}
\usepackage{graphicx}
\usepackage{fancyhdr}
\usepackage{booktabs}
\usepackage{amssymb}
\usepackage{subcaption}
\usepackage{tabularx}
\usepackage{adjustbox}
\usepackage{array}
\usepackage{url}
\usepackage{epstopdf}
\begin{document}
\title{GL-PGENet: A Parameterized Generation Framework for Robust Document Image Enhancement}
\author{Zhihong Tang  \\
\texttt{sichuantzh@gmail.com}}
\maketitle
\begin{abstract}
\begin{quote}
\textbf{D}ocument \textbf{I}mage \textbf{E}nhancement (DIE) serves as a critical component in Document AI systems, where its performance substantially determines the effectiveness of downstream tasks. To address the limitations of existing methods confined to single-degradation restoration or grayscale image processing, we present \textbf{G}lobal with \textbf{L}ocal \textbf{P}arametric \textbf{G}eneration \textbf{E}nhancement Network (GL-PGENet), a novel architecture designed for multi-degraded color document images, ensuring both efficiency and robustness in real-world scenarios. Our solution incorporates three key innovations: First, a hierarchical enhancement framework that integrates global appearance correction with local refinement, enabling coarse-to-fine quality improvement. Second, a Dual-Branch Local-Refine Network with parametric generation mechanisms that replaces conventional direct prediction, producing enhanced outputs through learned intermediate parametric representations rather than pixel-wise mapping. This approach enhances local consistency while improving model generalization. Finally,  a modified NestUNet architecture incorporating dense block to effectively fuse low-level pixel features and high-level semantic features, specifically adapted for document image characteristics. In addition, to enhance generalization performance, we adopt a two-stage training strategy: large-scale pretraining on a synthetic dataset of 500,000+ samples followed by task-specific fine-tuning. Extensive experiments demonstrate the superiority of GL-PGENet, achieving state-of-the-art SSIM scores of 0.7721 on DocUNet and 0.9480 on RealDAE. The model also exhibits remarkable cross-domain adaptability and maintains computational efficiency for high-resolution images without performance degradation, confirming its practical utility in real-world scenarios.
\end{quote}
\end{abstract}

\begin{figure*}[t]
    \centering
    \includegraphics[width=1\textwidth]{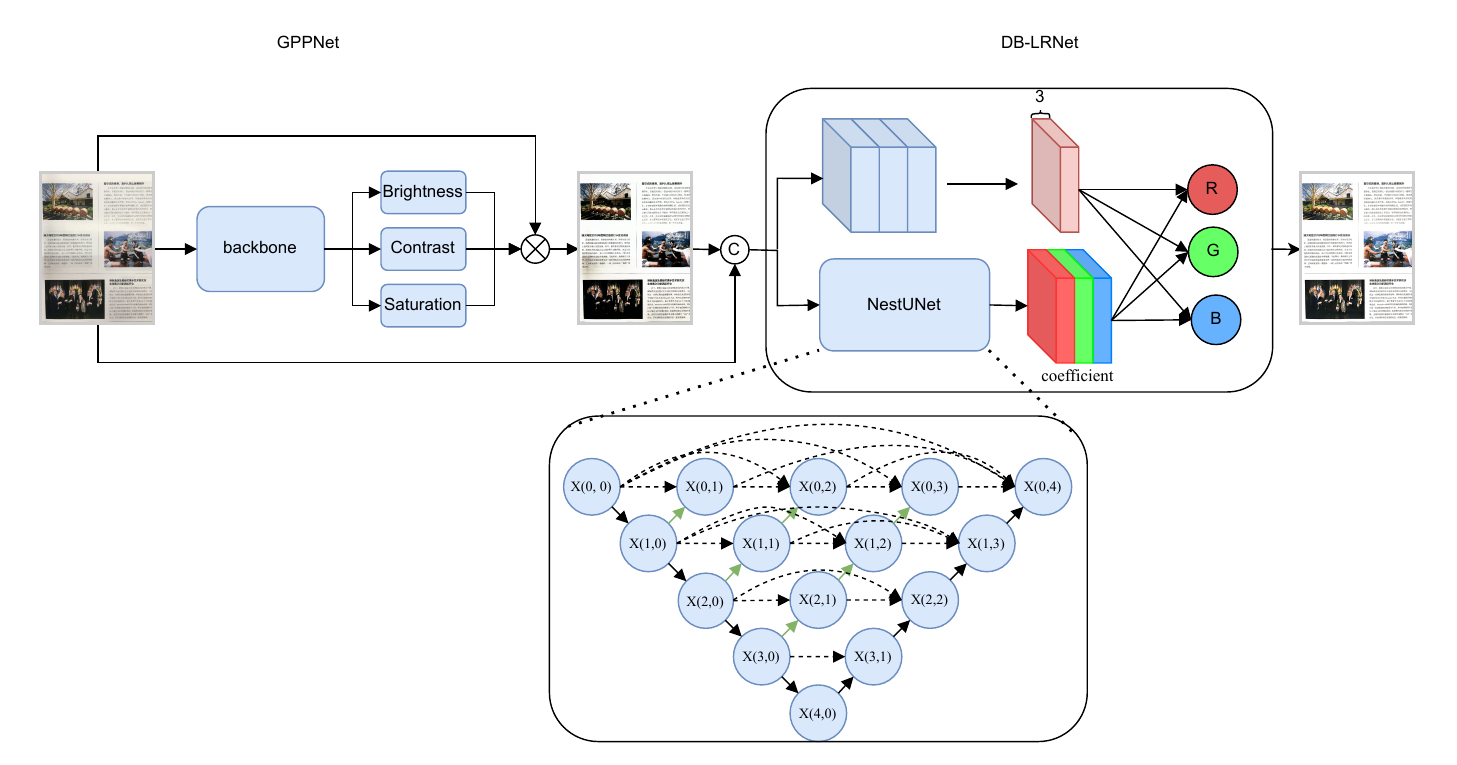}
    \caption{\textbf{Overview of the proposed GL-PGENet framework.} The architecture follows a coarse-to-fine two-stage paradigm. (a) The \textbf{GPPNet} first estimates global enhancement parameters for brightness, contrast, and saturation transformations to generate globally enhanced images with illumination consistency. (b) The \textbf{DB-LRNet} refines detail features: one branch employs convolutional operations for image smoothing, while the other utilizes a dense block-integrated NestUNet to learn linear transformation parameters. The final enhanced image is synthesized through the fusion of dual-branch outputs, achieving a balance between high-frequency detail preservation and local contextual consistency adaptation in color document image enhancement.}
    \label{fig:framework}
\end{figure*}

\section{Introduction}
The proliferation of smartphone and tablet technologies has revolutionized the digitization of printed materials, encompassing diverse media ranging from bound publications to ephemeral handwritten notes. Unlike conventional document scanning systems, the quality of document images captured by handheld mobile devices is often adversely affected by environmental and technical factors, including various lighting conditions, camera angles, and paper quality. These factors may cause the degradation of document images (e.g., shadows, blurs, wrinkles, and color casts, etc.), which negatively affect the readability and usability of digitized documents \cite{zamora2007behaviour}. Moreover, such deficiencies can also hinder the performance of document AI systems, particularly optical character recognition (OCR), which rely on high-quality images to produce accurate results \cite{souibgui2020gan,das2019dewarpnet}. To address these challenges, researchers have developed various techniques for DIE (Document Image Enhancement). 

Early document image enhancement (DIE) methodologies predominantly employed conventional image processing techniques, including histogram equalization and contrast stretching \cite{xiong2018degraded}. While effective for basic quality improvements, these approaches demonstrate limited capability in handling complex multi-degradation scenarios prevalently in real-world color document images. The advent of deep learning has catalyzed significant advancements in DIE through convolutional neural networks (CNNs) that learn nonlinear mappings between degraded inputs and enhanced outputs \cite{souibgui2020gan,zhang2023appearance,feng2021doctr}. Contemporary image restoration research predominantly focuses on specialized models targeting individual degradation types - exemplified by BEDSR-Net \cite{lin2020bedsr} for shadow removal and DeepDeblur \cite{mei2019deepdeblur} for deblurring.

Recent efforts to address multi-degradation challenges in color document images, such as the two-stage GCDRNet \cite{zhang2023appearance} and transformer-based DocStormer \cite{liu2023docstormerrevitalizingmultidegradedcolored}, reveal two critical limitations: (1) high computational complexity that impedes high-resolution processing, and (2) suboptimal enhancement quality for real-world color document images. While unsupervised approaches like UDoc-GAN \cite{wang2022udoc} attempt to mitigate data scarcity issues, their performance on color document image enhancement remains unsatisfactory. These challenges motivate our investigation into efficient multi-degradation enhancement solutions.

To address these critical gaps, we present GL-PGENet, a novel framework that achieves state-of-the-art performance in multi-degradation color document enhancement while maintaining computational efficiency. Our principal contributions include:
\begin{itemize}
    \item[\textbullet] A novel two-stage enhancement architecture featuring a streamlined global processing stage with parametric regression mechanism, significantly reducing computational overhead while preserving global consistency.
    \item[\textbullet] A Dual-Branch Local-Refine Network that innovatively predicts linear transformation parameters for final enhancement, departing from conventional direct prediction paradigms
    \item[\textbullet] The Dual-Branch Local-Refine Network incorporates a modified nested UNet \cite{zhou2018unet++} with dense blocks \cite{huang2017densely}, effectively preserving high-frequency details while enhancing semantic feature extraction.
    \item[\textbullet] Demonstrated computational scalability achieving about 75\% inference time reduction for high-resolution images without model retraining or significant quality degradation
\end{itemize}
The proposed GL-PGENet advances practical deployment of document enhancement systems through its balanced optimization of enhancement quality and computational efficiency. Our framework addresses the critical industry need for real-time processing of high-resolution color documents under complex degradation conditions.

\section{Related Works}
\subsection{Document Image Enhancement}
Document image enhancement play an important component in document analysis systems, forming the foundational stage of intelligent document processing pipelines to improve both readability and visual quality \cite{liu2023docstormerrevitalizingmultidegradedcolored}. Before the emergence of deep learning, traditional document enhancement methodologies primarily relied on thresholding-based approaches, achieving foreground-background separation through intensity differentiation for image quality improvement. The FAIR framework \cite{lelore2013fair} implements a double-threshold edge detection algorithm designed to balance detail preservation and noise suppression. Alternative methodologies include a fuzzy expert system integrated with adaptive pixel-counting mechanisms for global threshold optimization \cite{annabestani2019new}. And a spatial segmentation approach proposed by Xiong et al \cite{xiong2018degraded} partitions the document image into non-overlapping $w\times w$ sub-regions for localized contrast enhancement, followed by SVM-based threshold selection. However, empirical analyses reveal that these methods are highly sensitive to the document condition.

The evolution of deep learning has revolutionized document analysis research. In particularly, Generative Adversarial Networks (GANs) \cite{goodfellow2020generative} have demonstrated remarkable success in image generation and translation tasks \cite{isola2017image,zhang2023appearance}. Initial research focused on single-degradation scenarios. For shadow removal, BEDSR-Net \cite{lin2020bedsr} presents an innovative deep learning framework integrating a dedicated background estimation module that significantly enhances visual quality and text legibility through global background color. Zhang et al. \cite{Zhang_2023_CVPR} further advance this direction with a color-aware background extraction network (CBENet) and background-guided shadow removal network (BGShadowNet). Frequency domain decomposition approaches, such as those proposed by Li et al. \cite{Li_2023_ICCV}, enable effective learning of both low-level pixel features and high-level semantic features. In the document image deblurring, DeepDeblur \cite{mei2019deepdeblur} employs a 20-layer network with Sequential Highway Connections(SHC) to address convergence challenges. Recent work by DocDiff \cite{yang2023docdiff} presents the first diffusion-based framework for document deblurring, combining coarse low-frequency restoration with high-frequency residual refinement. DE-GAN \cite{9187695} demonstrates an end-to-end framework using conditional GANs for multi-task restoration including enhancement and watermark removal. However, these approaches demonstrate limited effectiveness in real-world scenarios characterized by concurrent degradations.

Recent efforts addressing multiple degradations in color documents remain limited. Both DocProj \cite{li2019document} and DocTr \cite{feng2021doctr} incorporate enhancement modules after geometric correction, but their effectiveness is constrained by synthetic training data with insufficient scale.  DocStormer \cite{liu2023docstormerrevitalizingmultidegradedcolored} introduces a Perceive-then-Restore paradigm with reinforced transformer blocks restoring multi-degration color document images to pristine PDF quality. Due to constraints in computational resources, the processing of large resolution images poses significant challenges. DocRes \cite{Zhang_2024_CVPR} proposes a novel multi-task restoration transformer \cite{vaswani2017attention} framework by taking different condition input, which is called Dynamic Task-Specific Prompt (DTSPrompt), a novel visual prompt approach that provides distinct prior features for different tasks. It's also implemented for the DIE task, but faces similar resource constraints. GCDRNet \cite{zhang2023appearance} advances multi-degradation color document image through a two stage architecture: GC-Net performs global contextual modeling and DR-Net conducts multi-scale detail restoration, integrated with a comprehensive multi-loss training strategy. While UDoc-GAN \cite{wang2022udoc} introduces unpaired training through modified cycle consistency constraints, though with limited success on color documents.

Current DIE methods exhibit critical limitations in preserving essential information when handling complex multi-degradation scenarios.  This challenge proves particularly acute in color document enhancement, where chromatic information carries significant semantic content. These limitations highlight the need for improved DIE algorithms to better address multi-degradation in real-world scenarios while maintaining critical document information.

\subsection{Generative Models}
Generative Adversarial Networks (GANs) \cite{goodfellow2020generative} represent a fundamental advancement in generative modeling through implicit density estimation. The GAN framework employs a dual-network architecture, comprising a generator and a discriminator, that is adversarially optimized based on game-theoretic principles. Initial implementations faced notable limitations in training stability and mode collapse phenomena, constraining output diversity and quality. To address these limitations, Wasserstein GANs \cite{arjovsky2017wassersteingan} were proposed, introducing the Wasserstein distance as an alternative loss function. Building on these advancements \cite{brock2018large,gulrajani2017improved}, subsequent research has significantly expanded the application of GANs across various domains, including image synthesis \cite{mirza2014conditional,karras2019style}, video generation \cite{tulyakov2018mocogan}, and domain adaptation \cite{9187695,isola2017image}.  

The recent emergence of diffusion models has significantly advanced generative model research. The Denoising Diffusion Probabilistic Model \cite{ho2020denoising} establishes a theoretical foundation by decomposing the image formation process through sequential denoising autoencoders guided by principles from non-quilibrium thermodynamics, achieving remarkable results in high-quality image synthesis. The framework was extended by Dhariwal and Nichol \cite{dhariwal2021diffusion} through classifier-guided sampling with Denoising Diffusion Implicit Models \cite{song2020denoising}. Despite these innovations demonstrate significant potential, conventional DM implementations face substantial computational challenges. To alleviate this limitation, Latent Diffusion Model \cite{rombach2022high} proposes an efficient adaptation by operating in the compressed latent space of pretrained autoencoders. This architectural innovation not only reduces computational overhead but also enhances model generalization capabilities. The evolution of diffusion architectures continues with Diffusion Transformers \cite{peebles2023scalable}, which replaces traditional U-Net structures. However, the inherent iterative sampling process remains a persistent challenge for real-world applications.

\subsection{Natural Image Enhancement}
Natural image enhancement can be divided into white-box and black-box methods. Here we mainly introduce white-box algorithm. For more information, please refer to the paper \cite{qi2021comprehensive}.
The Harmonizer \cite{ke2022harmonizer} formulates image enhancement as a parameter regression task for fundamental image filters, where neural networks predict interpretable transformation parameters (e.g., brightness, contrast) that are subsequently applied through image processing operations. The RSFNet \cite{ouyang2023rsfnet} is a white-box framework that employs parallel region-specific color filters to achieve fine-grained enhancements, mirroring professional colorists' divide-and-conquer methodology. Furthermore, Bianco et al. \cite{bianco2019learning} propose an efficient two-stage architecture that decouples parameter prediction from enhancement execution: transformation parameters are initially predicted from downsampled inputs, followed by full-resolution color adjustments, thereby achieving professional-grade retouching quality with optimized computational efficiency.

\subsection{Datasets}
\textbf{Real datasets}. Existing document enhancement datasets predominantly focus on single degradation types, such as the RDD dataset \cite{Zhang_2023_CVPR} for shadow removal and the TDD dataset \cite{9187695} for deblurring tasks. However, it is common that multiple degradations often occur together in real-world scenarios. The DocUNet dataset \cite{ma2018docunet} is a valuable resource for DIE, featuring a variety of degradations such as shadows, wrinkles, and bleed-through. But this dataset is limited by two key constraints: its small size(130 images) proves insufficient for model training, and its monolingual English documents restrict cross-linguistic applicability. The RealDAE dataset represents a high-quality collection of image pairs acquired through physical capture and professional post-processing. Specifically, the dataset was constructed by engaging expert photographers to meticulously retouch the raw degraded images using Adobe Photoshop\footnote{\url{https://www.adobe.com}} \cite{zhang2023appearance}. Meanwhile, we incorporate an additional dataset from Baidu AI Studio\footnote{\url{https://aistudio.baidu.com/datasetdetail/126294}} for generalization evaluation. Notably, this collection extends beyond the aforementioned degradation patterns by incorporating moire degradation, while primarily consisting of Chinese-language document samples.  Particularly, the dataset encompasses a diverse range of degradation types, offering comprehensive coverage of real-world document degradation scenarios. The availability of real-world datasets for this task remains significantly limited. To address this data scarcity, we further augment training through binary document datasets from the (H)-DIBCO series \cite{gatos2009icdar,ntirogiannis2014icfhr2014,pratikakis2010h,pratikakis2017icdar2017}.

\textbf{Synthetic datasets}
To address the aforementioned limitations of real-world datasets, researchers \cite{lin2020bedsr,das2020intrinsic} have employed rendering engines such as Blender\footnote{\url{https://www.blender.org}} to generate synthetic datasets. These datasets offer enhanced diversity, larger scale, and more accurate ground truths. It can simulate complex lighting conditions, paper shapes, environmental backgrounds, and camera positions. Despite recent advancements, the diversity and authenticity of synthetic datasets require further improvement. The Doc3D \cite{das2019dewarpnet} and Doc3dShade \cite{das2020intrinsic} demonstrate enhanced authenticity and degradation diversity compared to earlier benchmarks like DocProj \cite{li2019document}, they are primarily optimized for predicting low-frequency components. Moreover, the complex synthetic pipelines employed in these datasets pose significant implementation challenges, particularly in case-specific scenarios. To address these limitations, we implement a streamlined degradation synthesis pipeline using the Augraphy framework\footnote{\url{https://github.com/sparkfish/augraphy}} during pre-training. Empirical analysis reveals a non-monotonic correlation between degradation intensity and model efficacy, with optimal performance achieved at intermediate intensity levels.

\begin{figure}[h]
  \centering
  \begin{subfigure}[b]{0.11\textwidth}
    \includegraphics[width=\textwidth]{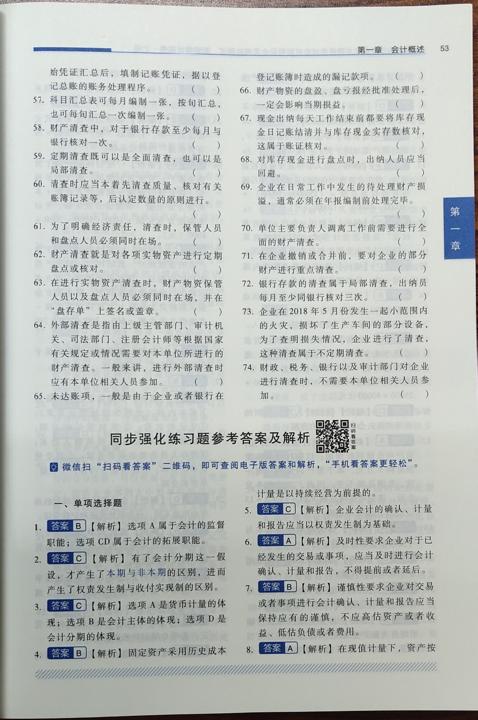}
  \end{subfigure}
  \hfill
  \begin{subfigure}[b]{0.11\textwidth}
    \includegraphics[width=\textwidth]{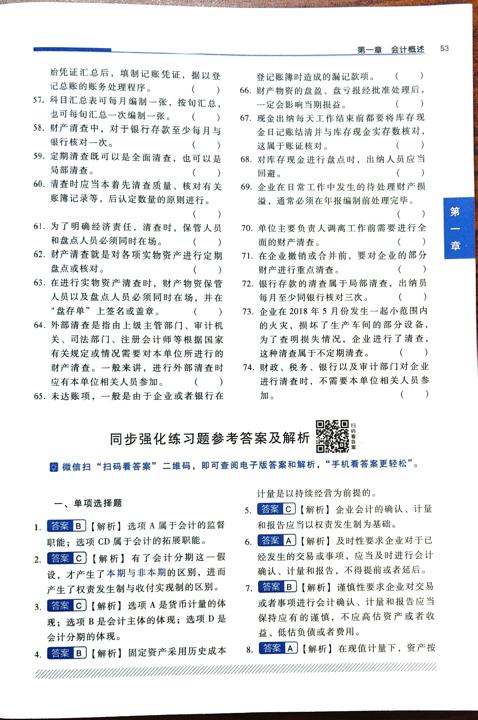}
  \end{subfigure}
  \hfill
  \begin{subfigure}[b]{0.11\textwidth}
    \includegraphics[width=\textwidth]{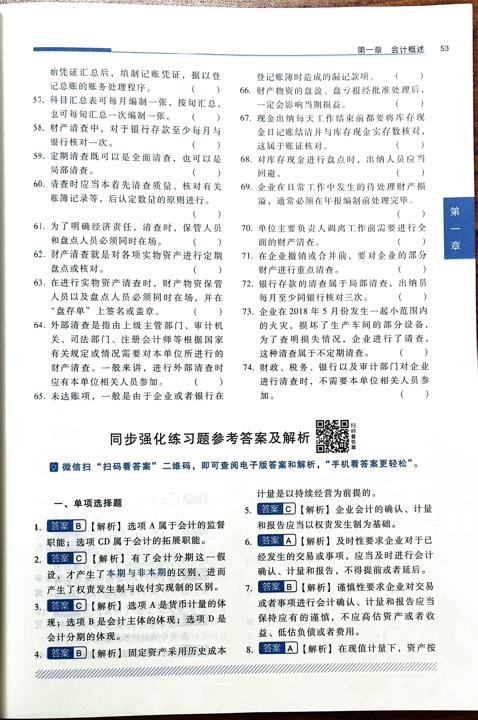}
  \end{subfigure}
  \hfill
  \begin{subfigure}[b]{0.11\textwidth}
    \includegraphics[width=\textwidth]{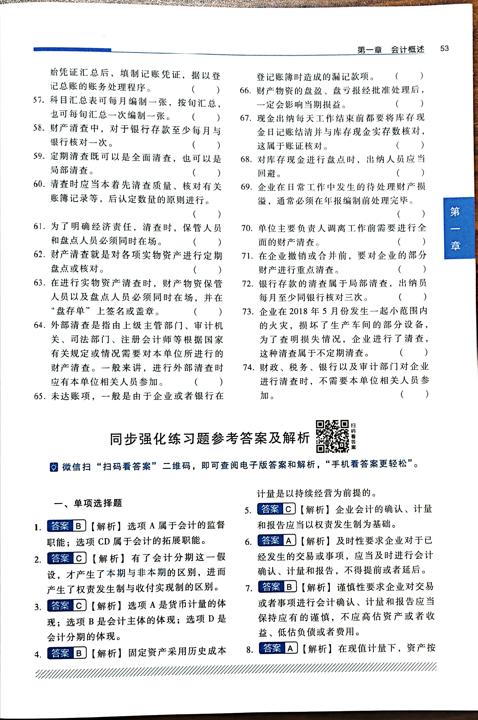}
  \end{subfigure}
  
  \begin{subfigure}[b]{0.11\textwidth}
    \includegraphics[width=\textwidth]{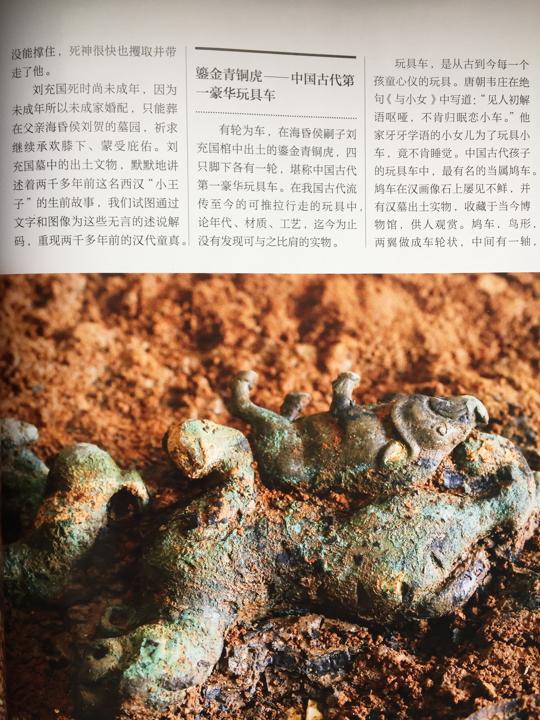}
  \end{subfigure}
  \hfill
  \begin{subfigure}[b]{0.11\textwidth}
    \includegraphics[width=\textwidth]{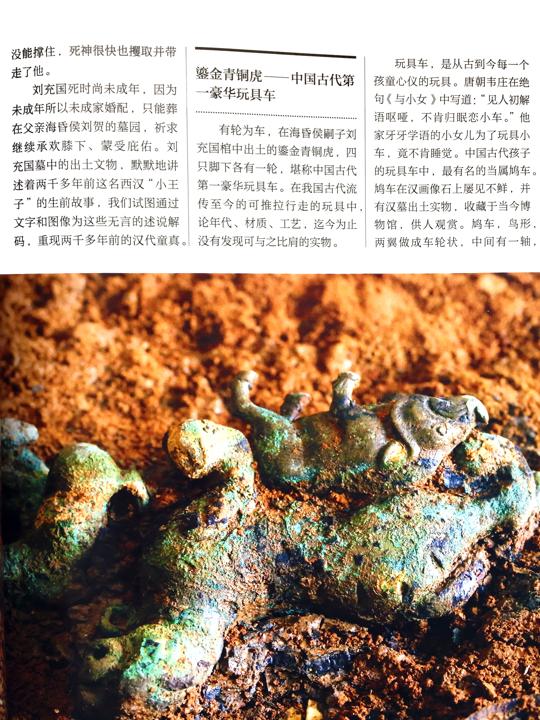}
  \end{subfigure}
  \hfill
  \begin{subfigure}[b]{0.11\textwidth}
    \includegraphics[width=\textwidth]{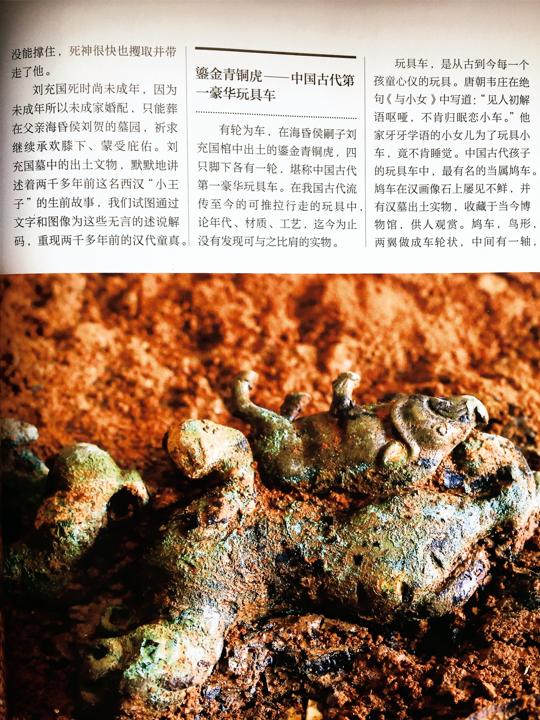}
  \end{subfigure}
  \hfill
  \begin{subfigure}[b]{0.11\textwidth}
    \includegraphics[width=\textwidth]{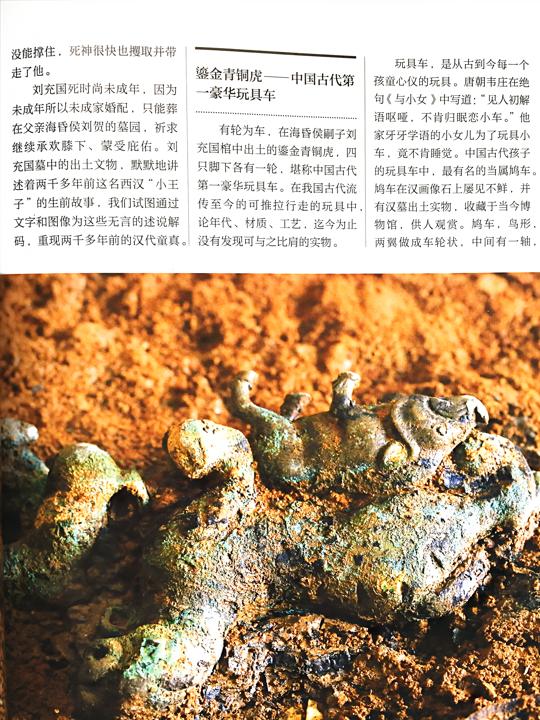}
  \end{subfigure}
  
  \begin{subfigure}[b]{0.11\textwidth}
    \includegraphics[width=\textwidth]{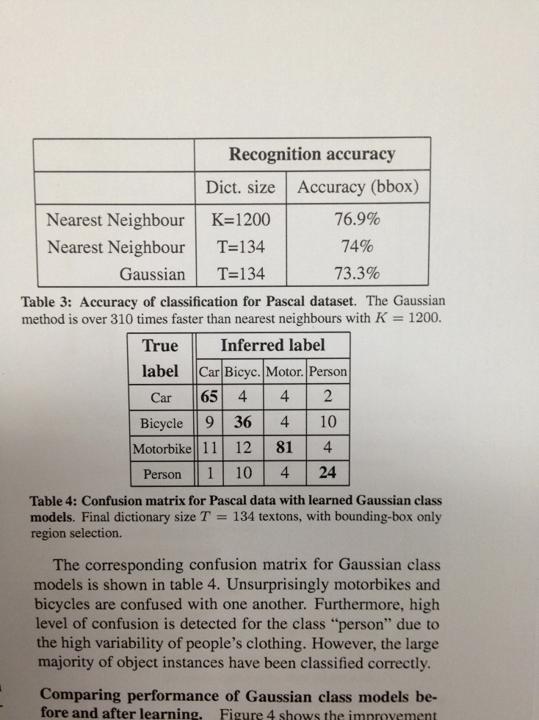}
    \caption{source}
  \end{subfigure}
  \hfill
  \begin{subfigure}[b]{0.11\textwidth}
    \includegraphics[width=\textwidth]{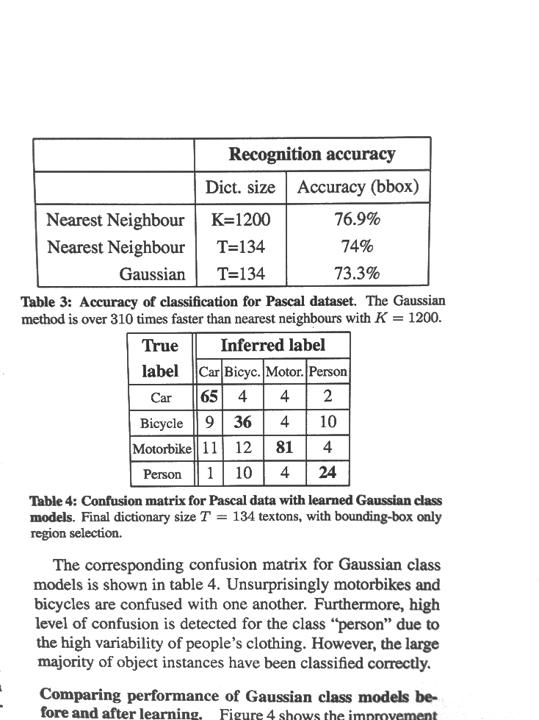}
    \caption{gt}
  \end{subfigure}
  \hfill
  \begin{subfigure}[b]{0.11\textwidth}
    \includegraphics[width=\textwidth]{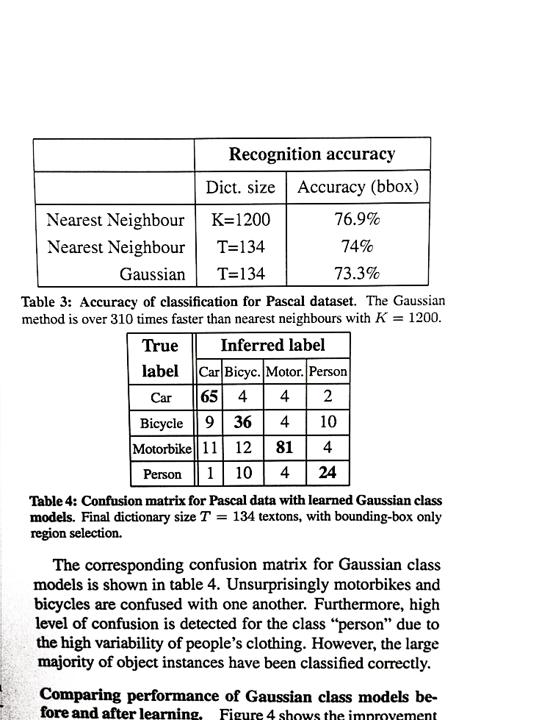}
    \caption{$I_g$}
  \end{subfigure}
  \hfill
  \begin{subfigure}[b]{0.11\textwidth}
    \includegraphics[width=\textwidth]{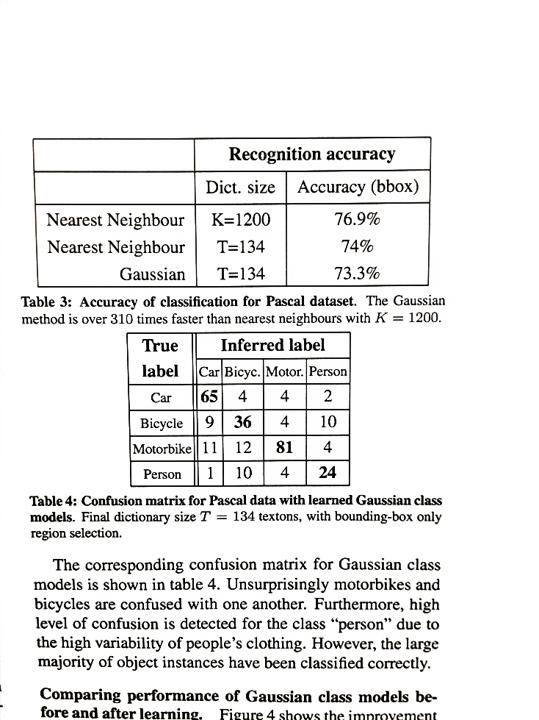}
    \caption{$I_e$}
  \end{subfigure}
  
 \caption{\textbf{Two-Stage Image Enhancement Process Visualization}. (a) Original degraded images; (b) Ground-truth reference images; (c) Global enhancement results $I_g$ from GPPNet with optimized brightness, contrast, and saturation parameters; (d) Final refined outputs $I_e$ generated by DB-LRNet demonstrating preserved high-frequency details alongside illumination consistency.}
  \label{fig:inter}
\end{figure}

\begin{figure*}[!t]
  \centering
  \begin{subfigure}[b]{0.12\textwidth}
    \includegraphics[width=\textwidth]{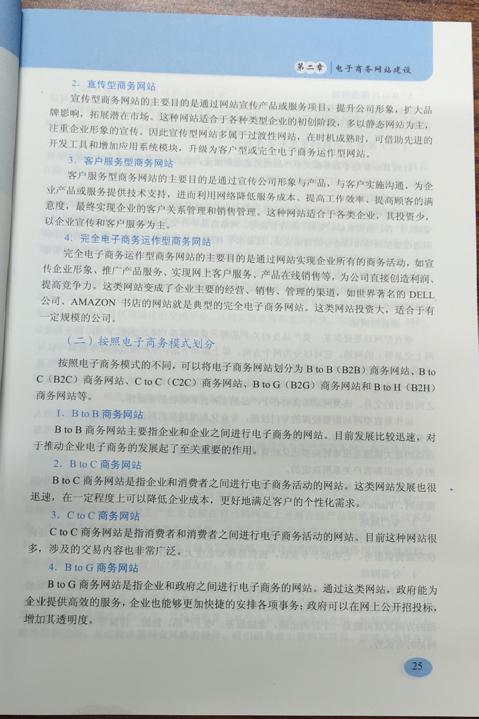}
  \end{subfigure}
  \hfill
  \begin{subfigure}[b]{0.12\textwidth}
    \includegraphics[width=\textwidth]{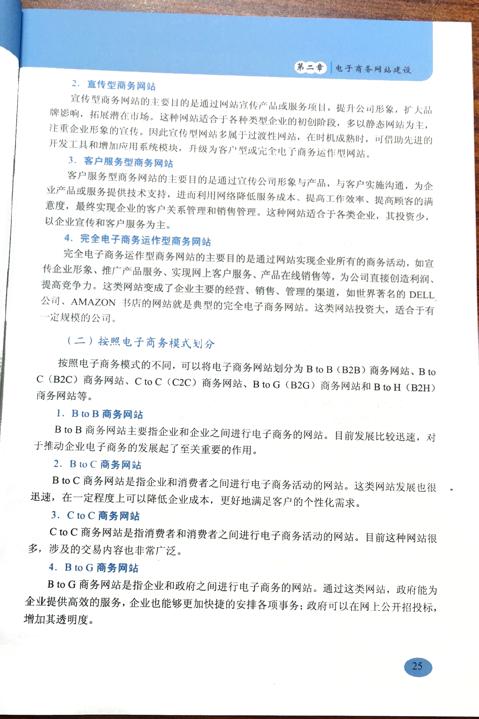}
  \end{subfigure}
  \hfill
  \begin{subfigure}[b]{0.12\textwidth}
    \includegraphics[width=\textwidth]{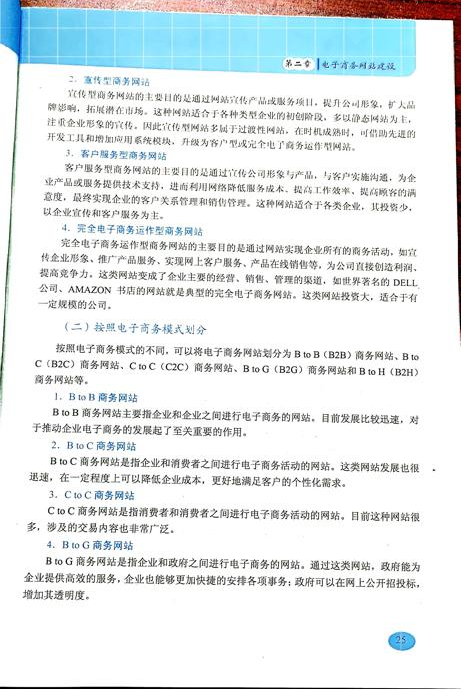}
  \end{subfigure}
  \hfill
  \begin{subfigure}[b]{0.12\textwidth}
    \includegraphics[width=\textwidth]{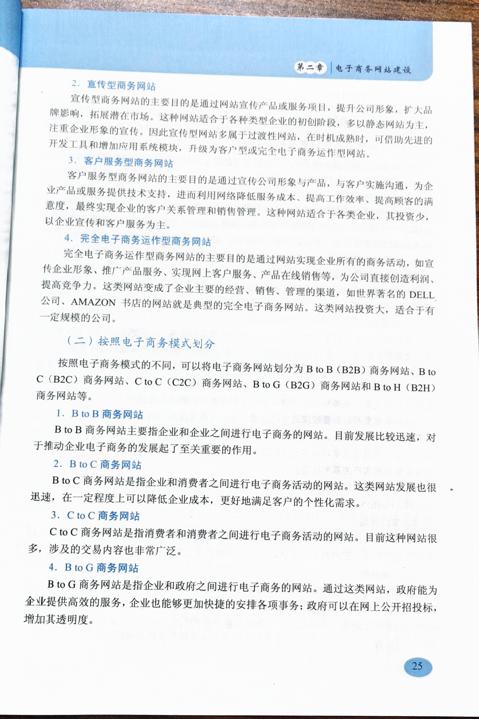}
  \end{subfigure}
  \begin{subfigure}[b]{0.12\textwidth}
    \includegraphics[width=\textwidth]{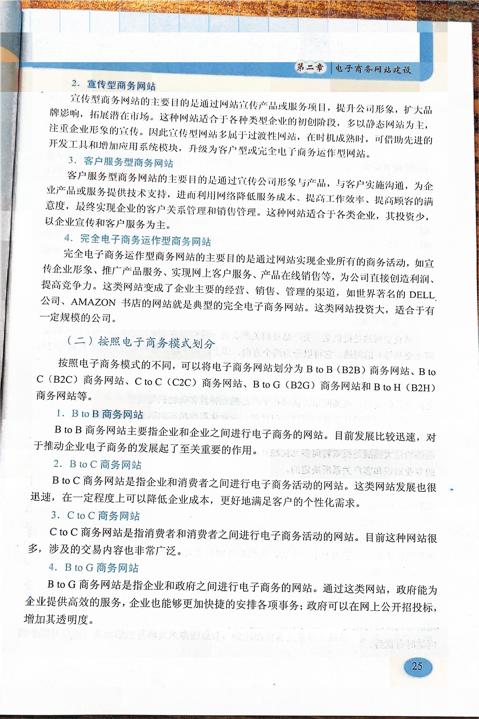}
  \end{subfigure}
  \hfill
  \begin{subfigure}[b]{0.12\textwidth}
    \includegraphics[width=\textwidth]{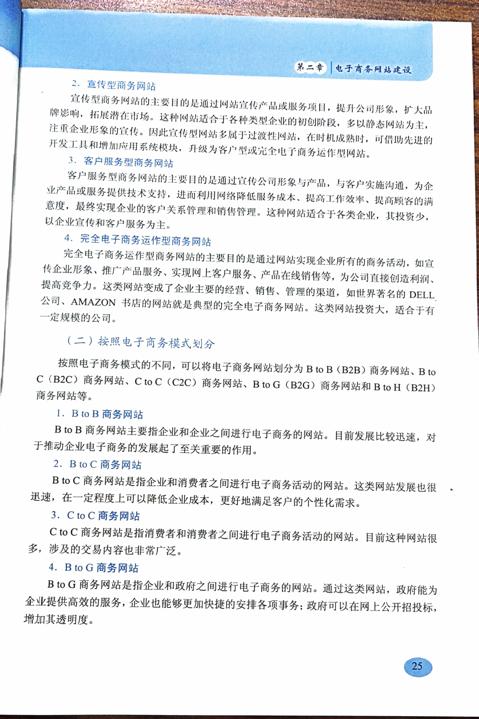}
  \end{subfigure}
  \hfill
  \begin{subfigure}[b]{0.12\textwidth}
    \includegraphics[width=\textwidth]{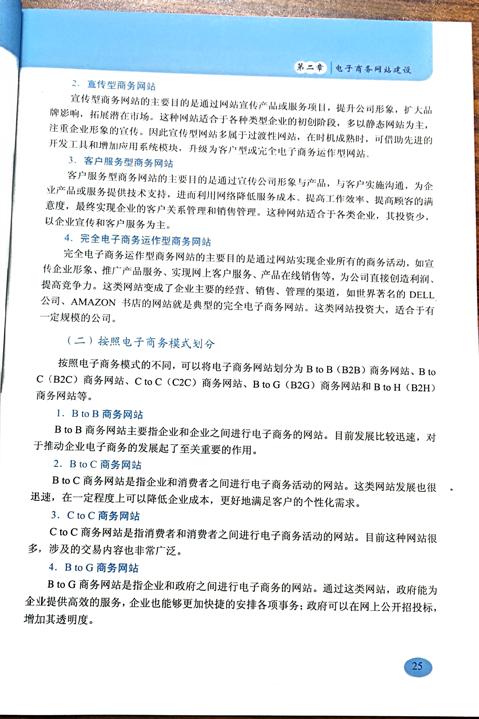}
  \end{subfigure}

  \begin{subfigure}[b]{0.12\textwidth}
    \includegraphics[width=\textwidth]{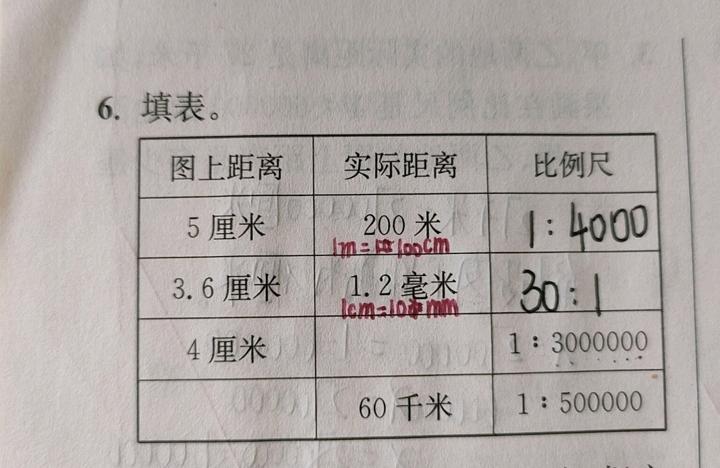}
  \end{subfigure}
  \hfill
  \begin{subfigure}[b]{0.12\textwidth}
    \includegraphics[width=\textwidth]{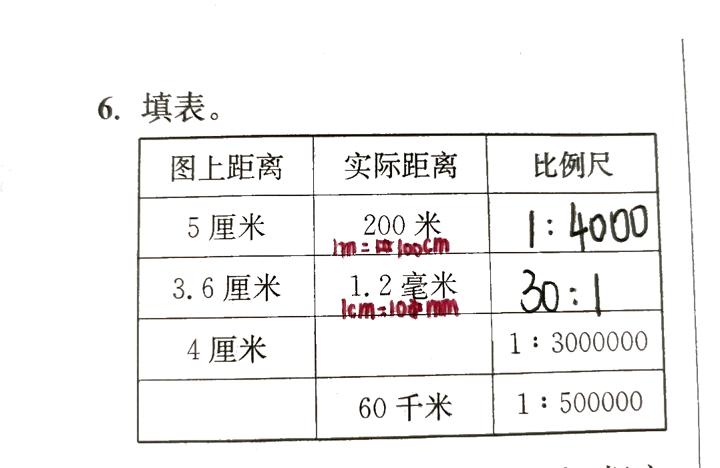}
  \end{subfigure}
  \hfill
  \begin{subfigure}[b]{0.12\textwidth}
    \includegraphics[width=\textwidth]{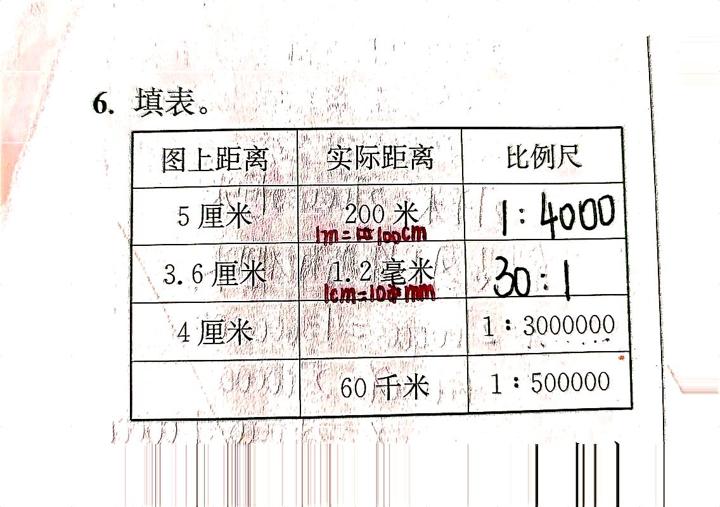}
  \end{subfigure}
  \hfill
  \begin{subfigure}[b]{0.12\textwidth}
    \includegraphics[width=\textwidth]{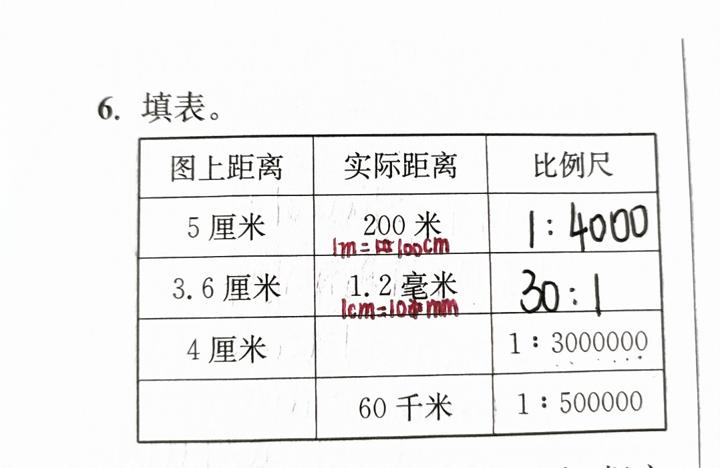}
  \end{subfigure}
  \begin{subfigure}[b]{0.12\textwidth}
    \includegraphics[width=\textwidth]{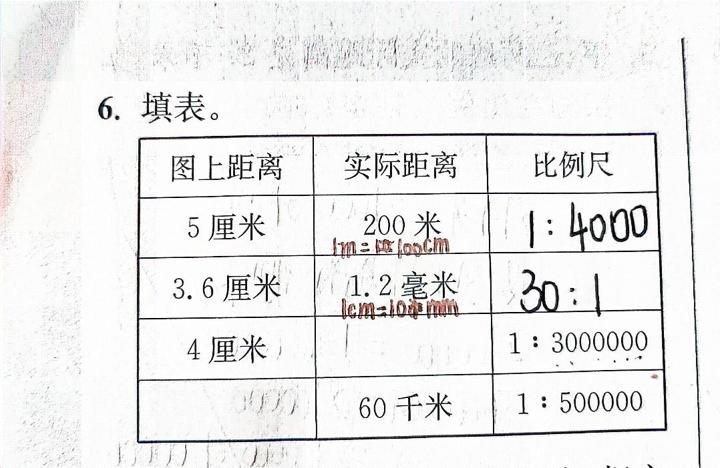}
  \end{subfigure}
  \hfill
  \begin{subfigure}[b]{0.12\textwidth}
    \includegraphics[width=\textwidth]{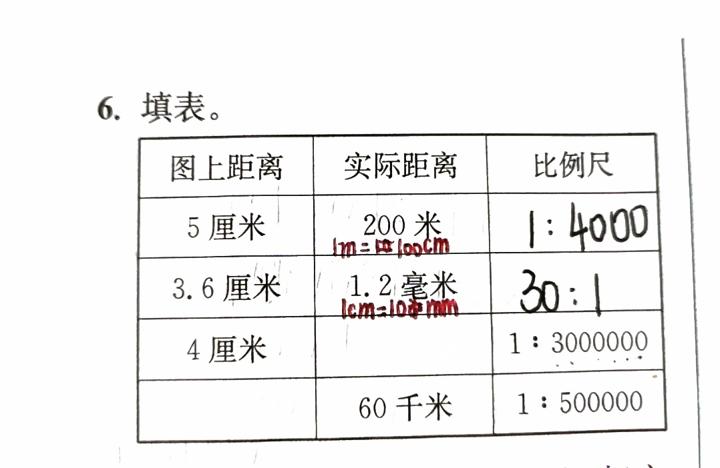}
  \end{subfigure}
  \hfill
  \begin{subfigure}[b]{0.12\textwidth}
    \includegraphics[width=\textwidth]{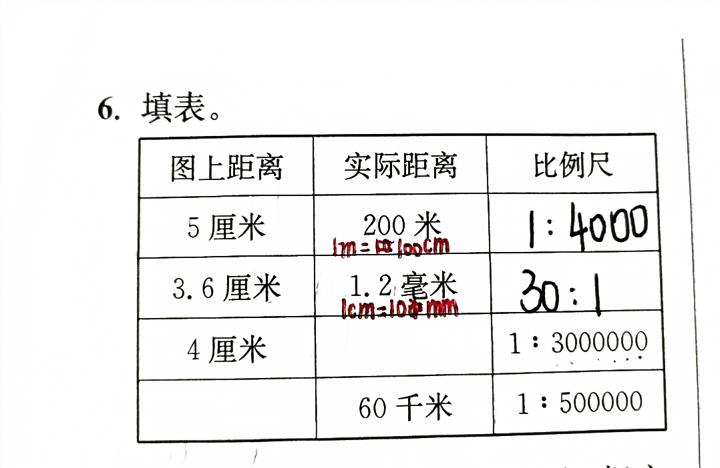}
  \end{subfigure}

  \begin{subfigure}[b]{0.12\textwidth}
    \includegraphics[width=\textwidth]{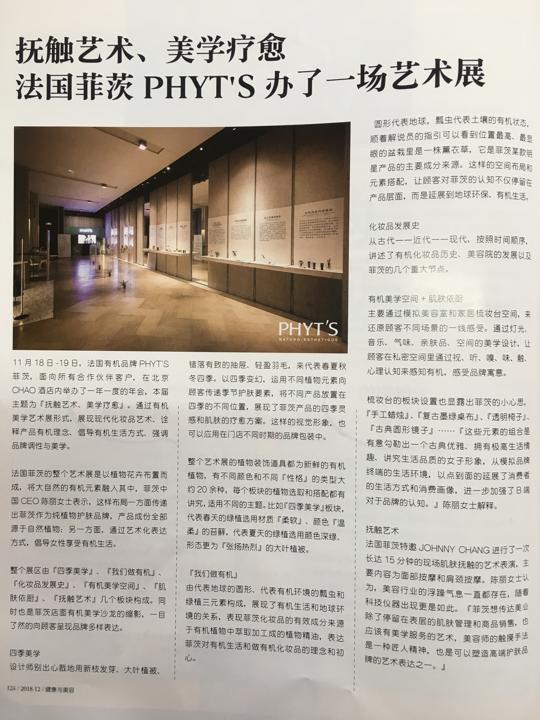}
  \end{subfigure}
  \hfill
  \begin{subfigure}[b]{0.12\textwidth}
    \includegraphics[width=\textwidth]{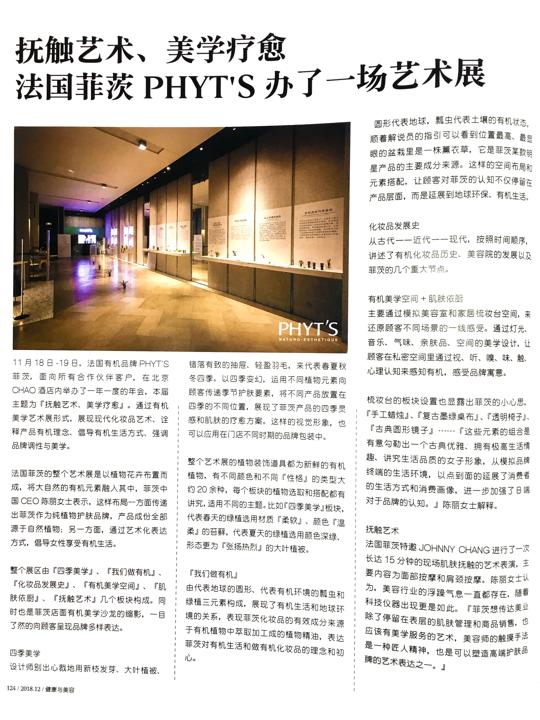}
  \end{subfigure}
  \hfill
  \begin{subfigure}[b]{0.12\textwidth}
    \includegraphics[width=\textwidth]{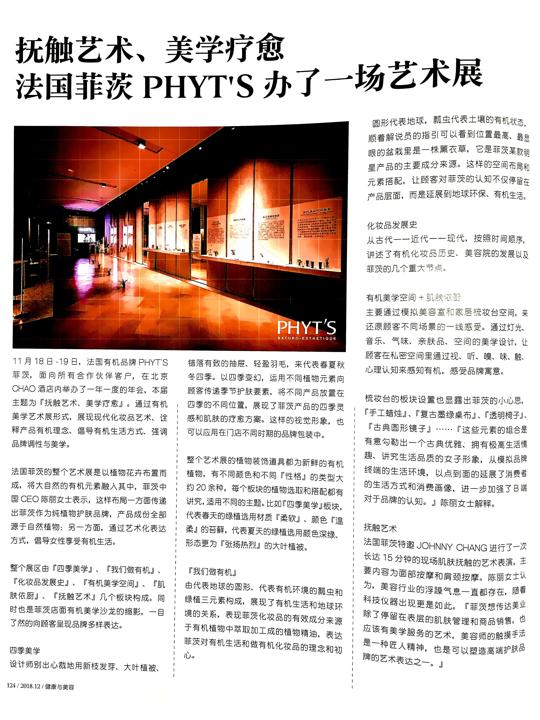}
  \end{subfigure}
  \hfill
  \begin{subfigure}[b]{0.12\textwidth}
    \includegraphics[width=\textwidth]{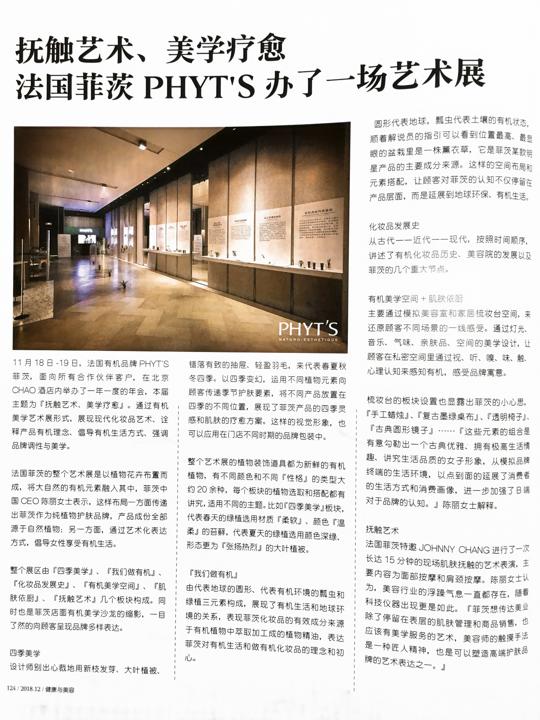}
  \end{subfigure}
  \begin{subfigure}[b]{0.12\textwidth}
    \includegraphics[width=\textwidth]{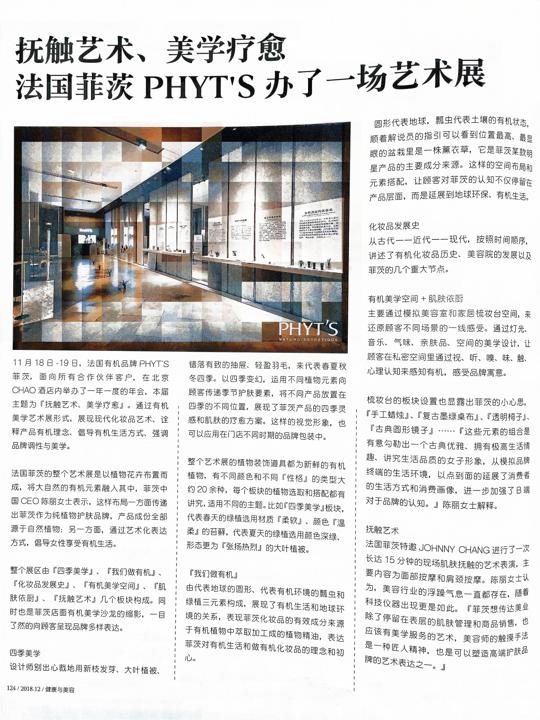}
  \end{subfigure}
  \hfill
  \begin{subfigure}[b]{0.12\textwidth}
    \includegraphics[width=\textwidth]{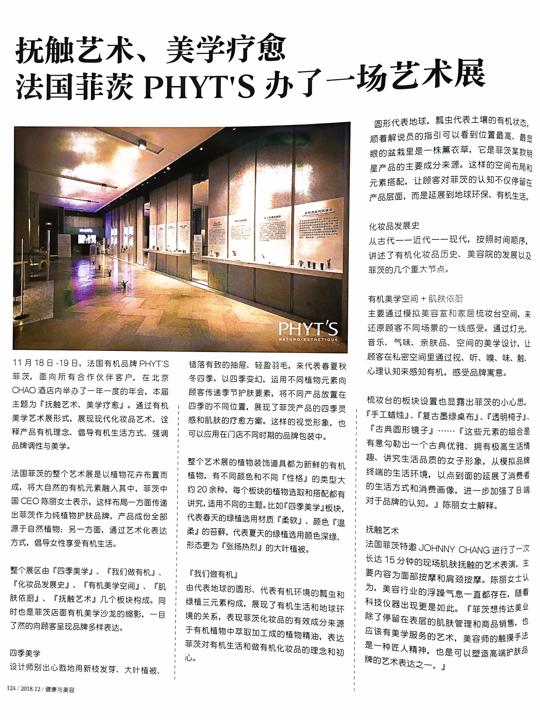}
  \end{subfigure}
  \hfill
  \begin{subfigure}[b]{0.12\textwidth}
    \includegraphics[width=\textwidth]{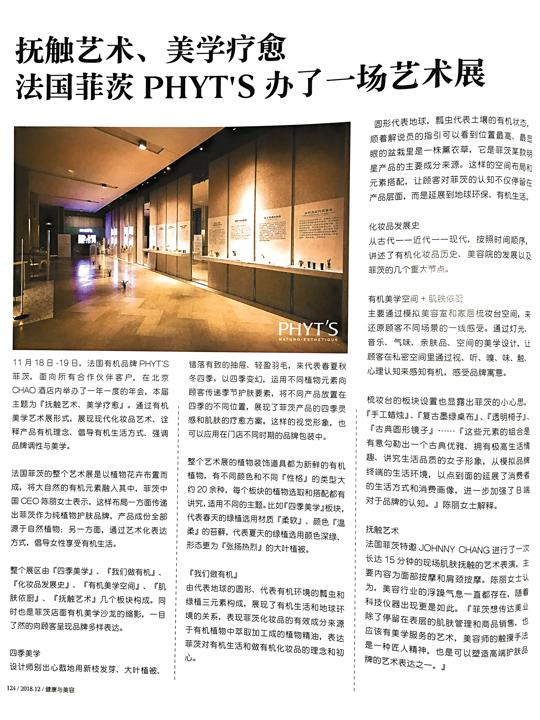}
  \end{subfigure}

  \begin{subfigure}[b]{0.12\textwidth}
    \includegraphics[width=\textwidth]{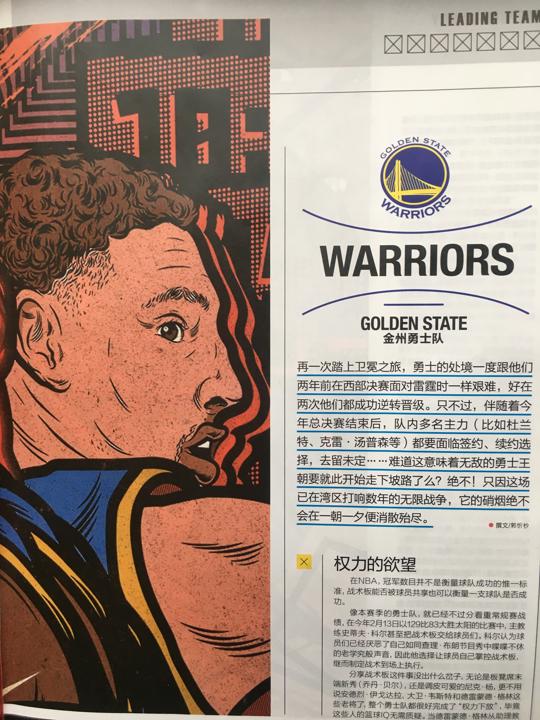}
  \end{subfigure}
  \hfill
  \begin{subfigure}[b]{0.12\textwidth}
    \includegraphics[width=\textwidth]{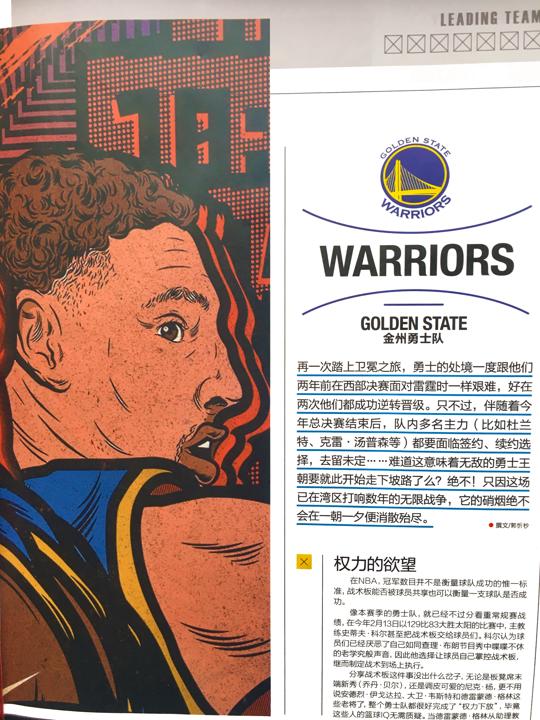}
  \end{subfigure}
  \hfill
  \begin{subfigure}[b]{0.12\textwidth}
    \includegraphics[width=\textwidth]{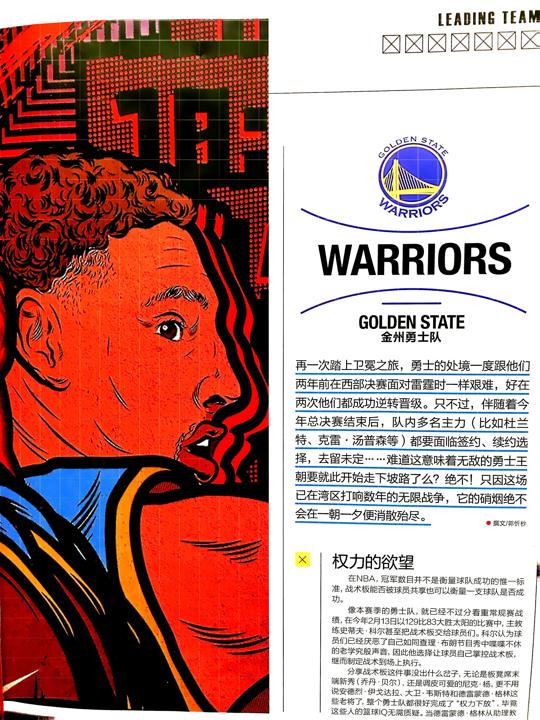}
  \end{subfigure}
  \hfill
  \begin{subfigure}[b]{0.12\textwidth}
    \includegraphics[width=\textwidth]{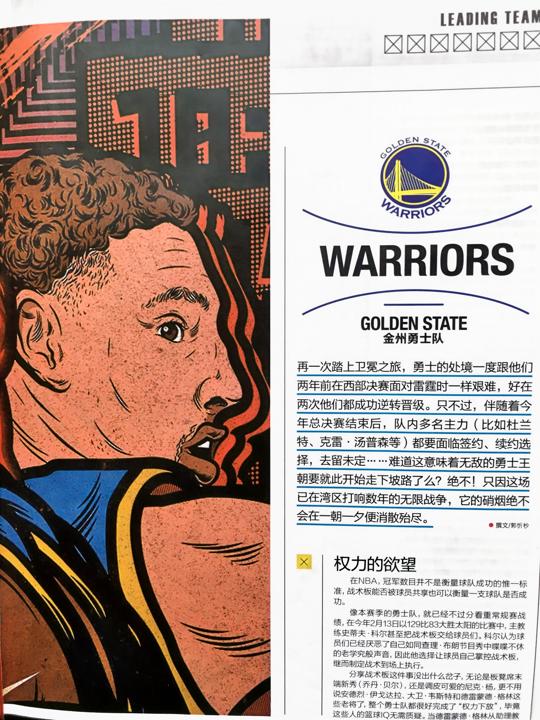}
  \end{subfigure}
  \begin{subfigure}[b]{0.12\textwidth}
    \includegraphics[width=\textwidth]{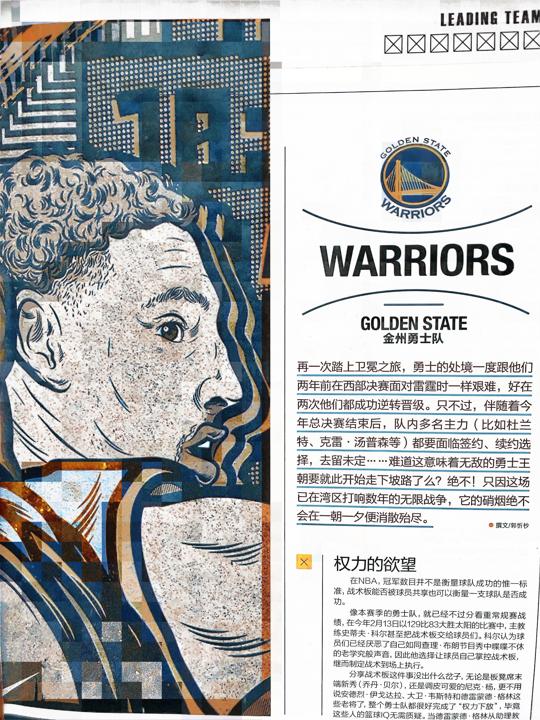}
  \end{subfigure}
  \hfill
  \begin{subfigure}[b]{0.12\textwidth}
    \includegraphics[width=\textwidth]{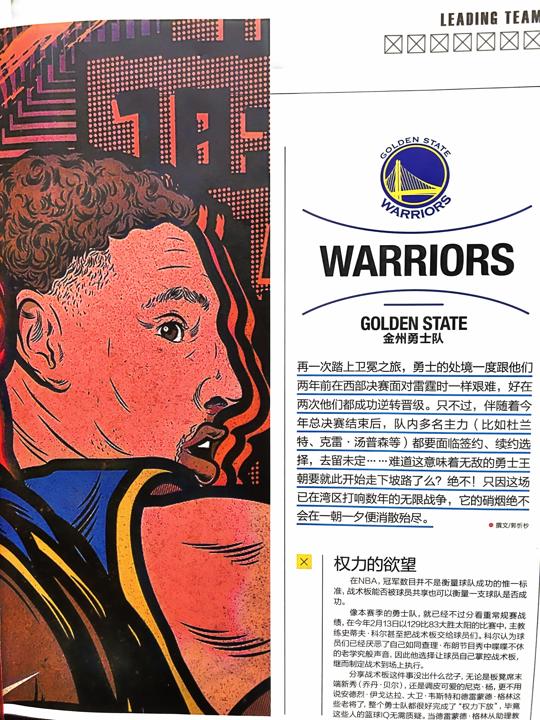}
  \end{subfigure}
  \hfill
  \begin{subfigure}[b]{0.12\textwidth}
    \includegraphics[width=\textwidth]{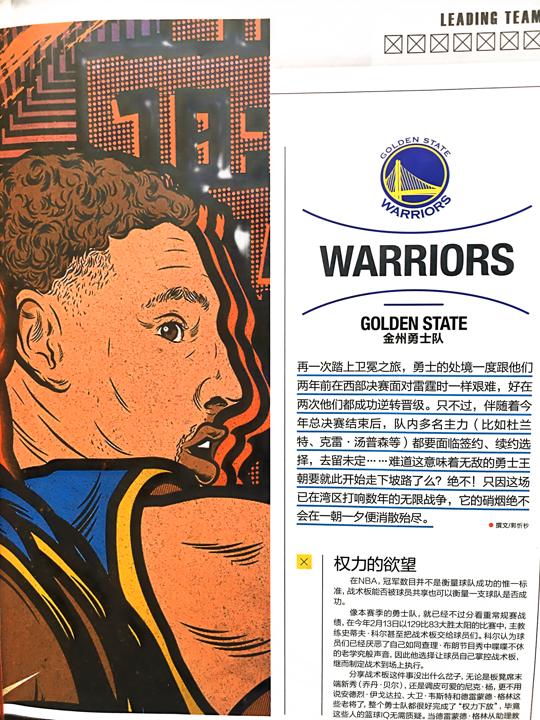}
  \end{subfigure}

  \begin{subfigure}[b]{0.12\textwidth}
    \includegraphics[width=\textwidth]{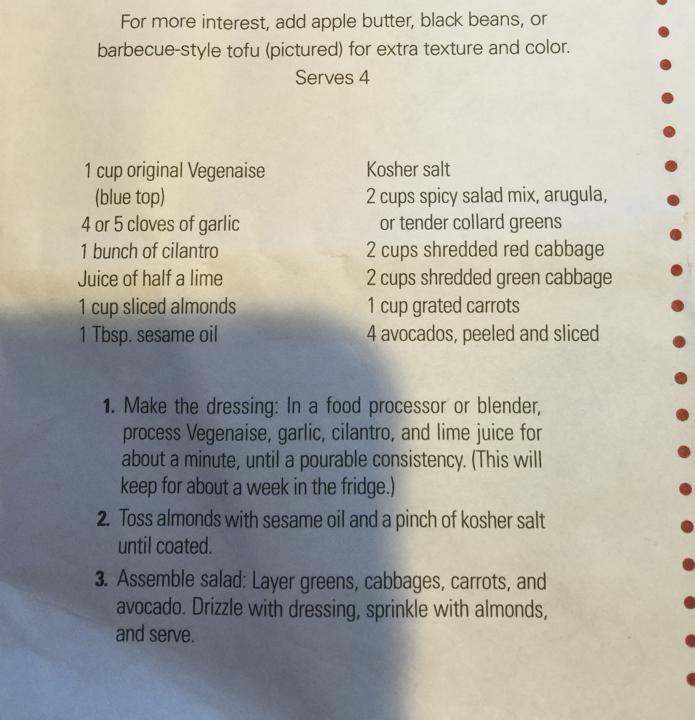}
  \end{subfigure}
  \hfill
  \begin{subfigure}[b]{0.12\textwidth}
    \includegraphics[width=\textwidth]{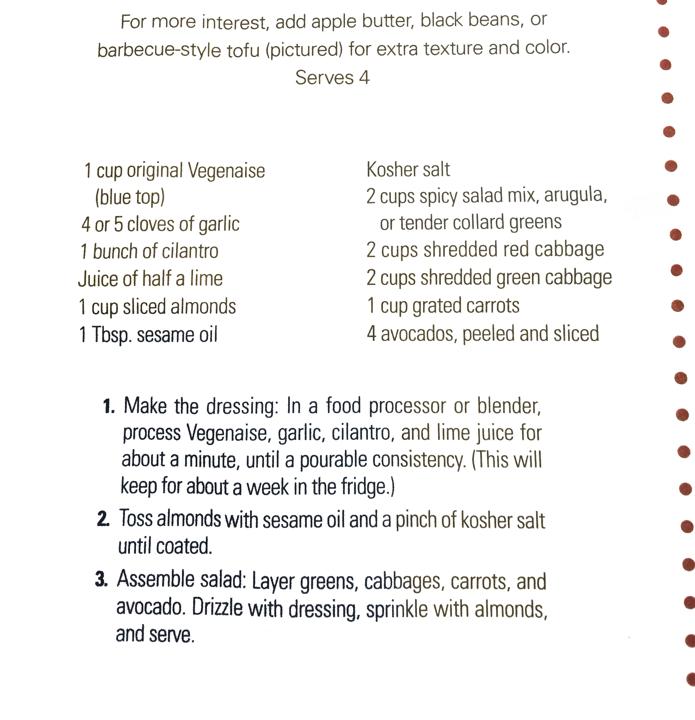}
  \end{subfigure}
  \hfill
  \begin{subfigure}[b]{0.12\textwidth}
    \includegraphics[width=\textwidth]{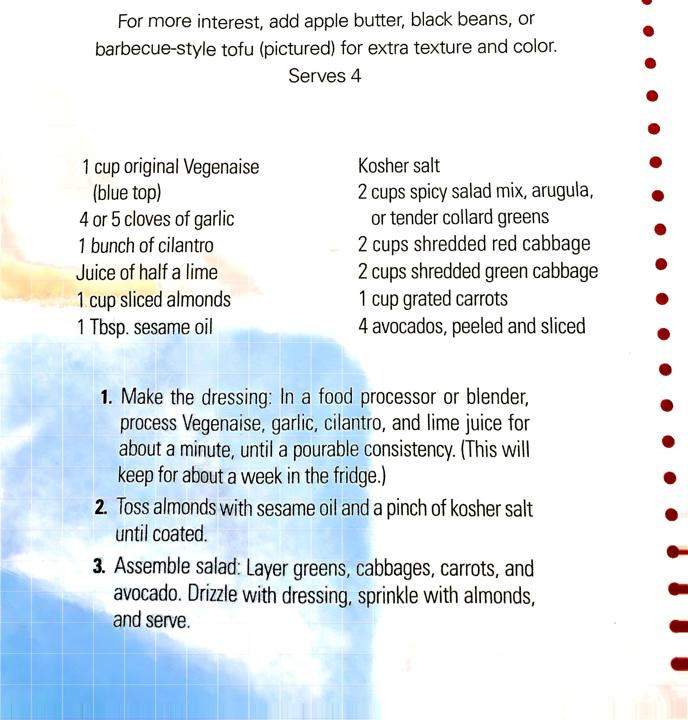}
  \end{subfigure}
  \hfill
  \begin{subfigure}[b]{0.12\textwidth}
    \includegraphics[width=\textwidth]{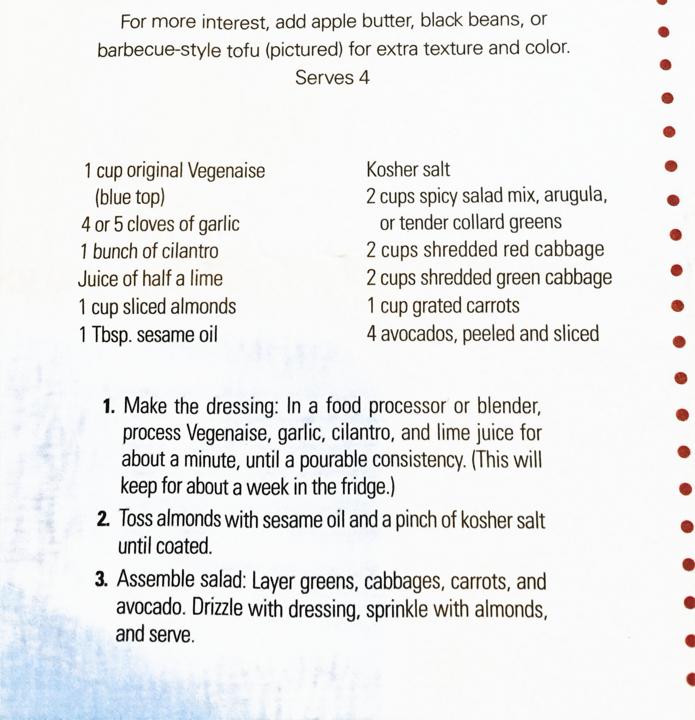}
  \end{subfigure}
  \begin{subfigure}[b]{0.12\textwidth}
    \includegraphics[width=\textwidth]{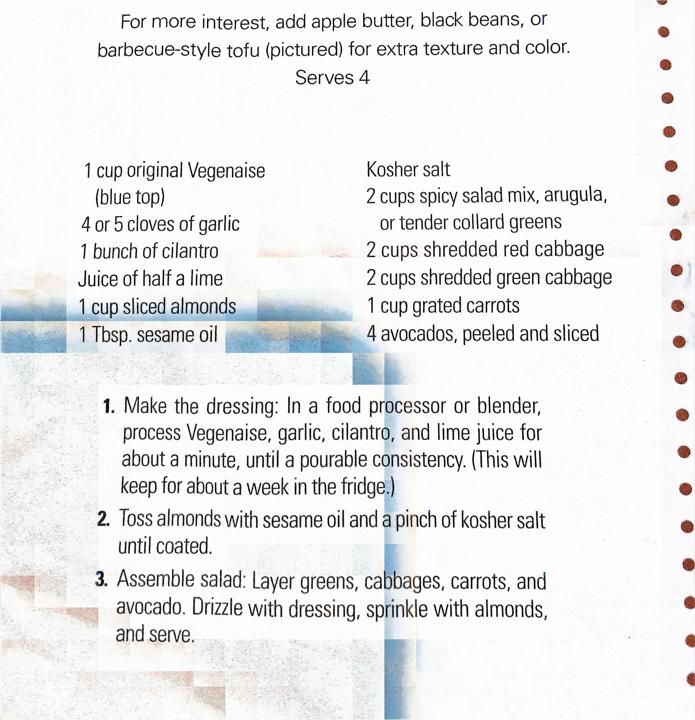}
  \end{subfigure}
  \hfill
  \begin{subfigure}[b]{0.12\textwidth}
    \includegraphics[width=\textwidth]{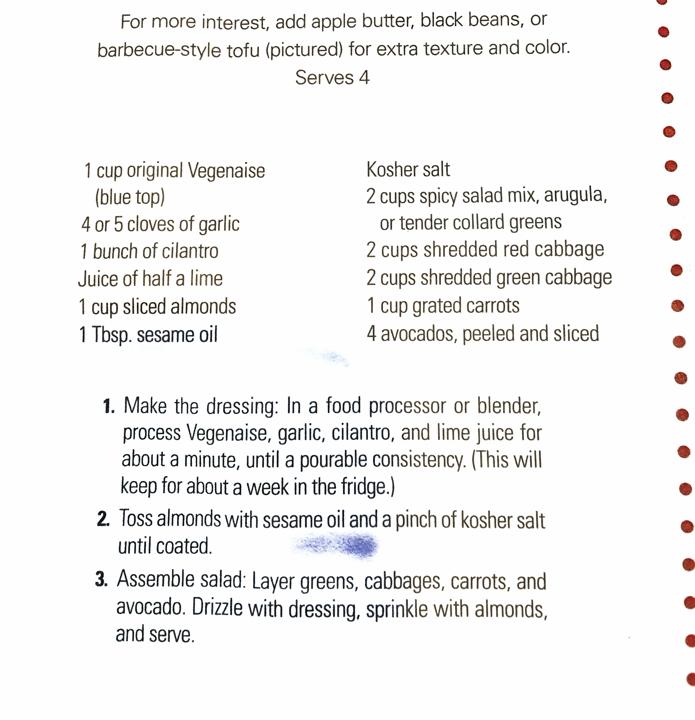}
  \end{subfigure}
  \hfill
  \begin{subfigure}[b]{0.12\textwidth}
    \includegraphics[width=\textwidth]{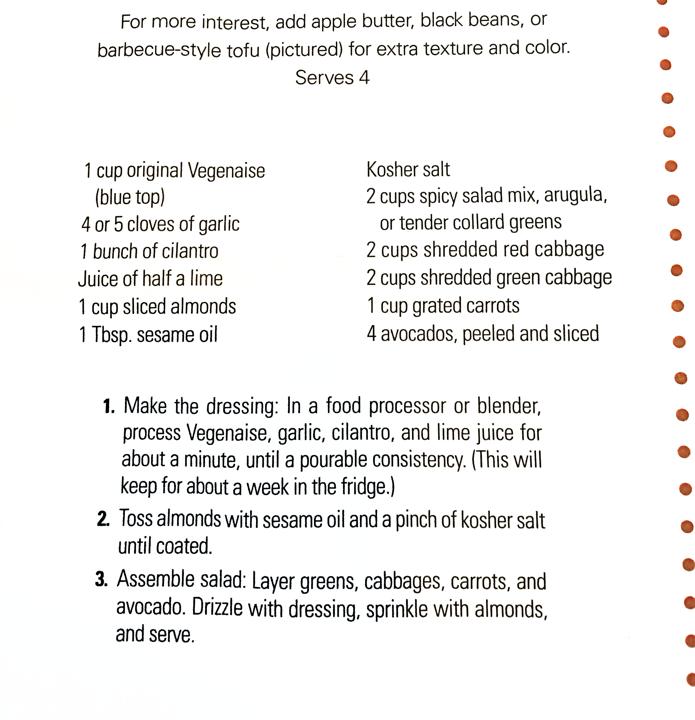}
  \end{subfigure}

  \begin{subfigure}[b]{0.12\textwidth}
    \includegraphics[width=\textwidth]{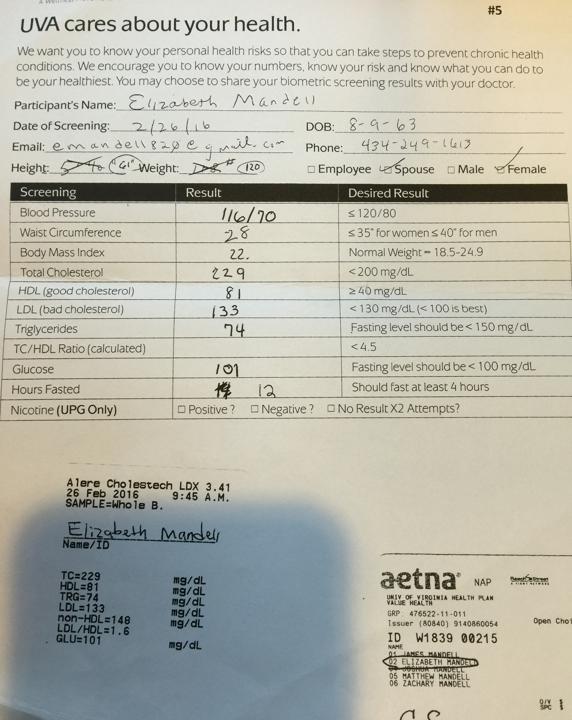}
    \caption{}
  \end{subfigure}
  \hfill
  \begin{subfigure}[b]{0.12\textwidth}
    \includegraphics[width=\textwidth]{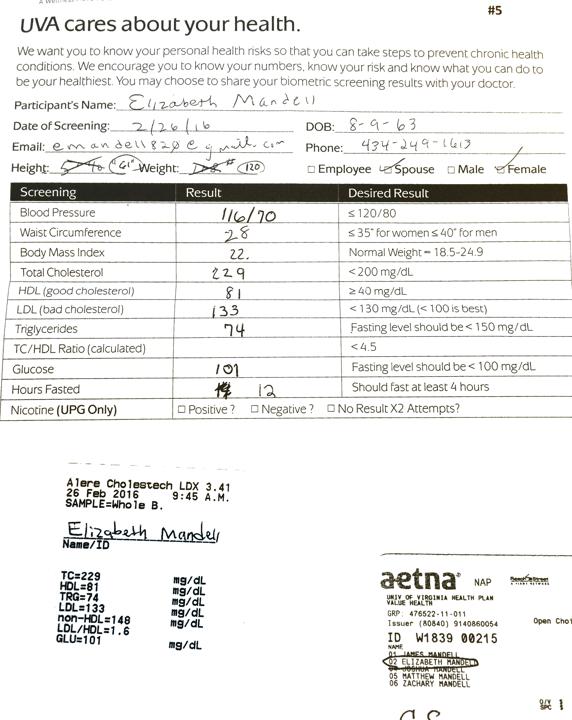}
    \caption{}
  \end{subfigure}
  \hfill
  \begin{subfigure}[b]{0.12\textwidth}
    \includegraphics[width=\textwidth]{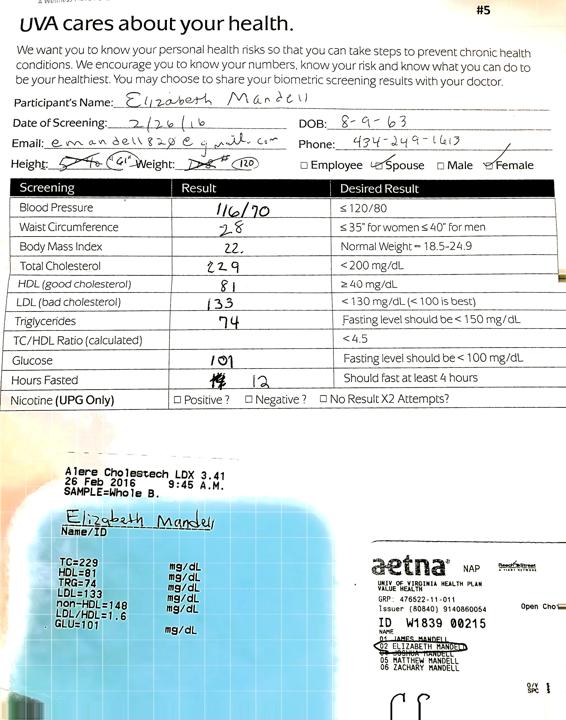}
    \caption{}
  \end{subfigure}
  \hfill
  \begin{subfigure}[b]{0.12\textwidth}
    \includegraphics[width=\textwidth]{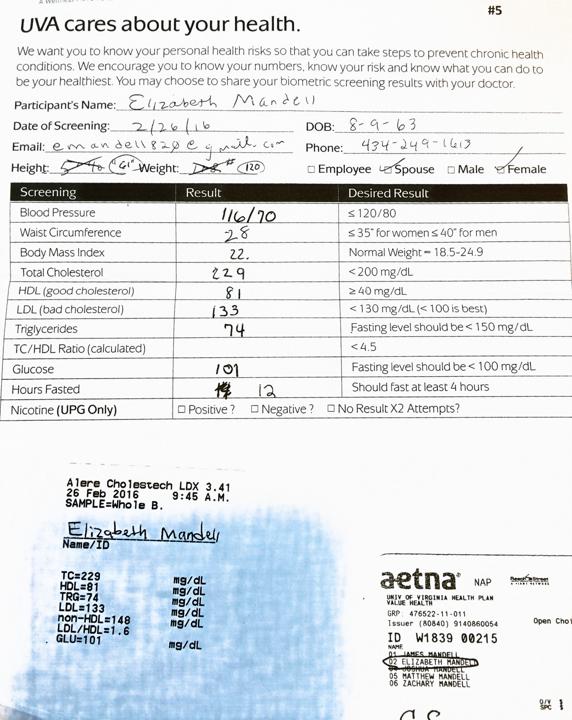}
    \caption{}
  \end{subfigure}
  \begin{subfigure}[b]{0.12\textwidth}
    \includegraphics[width=\textwidth]{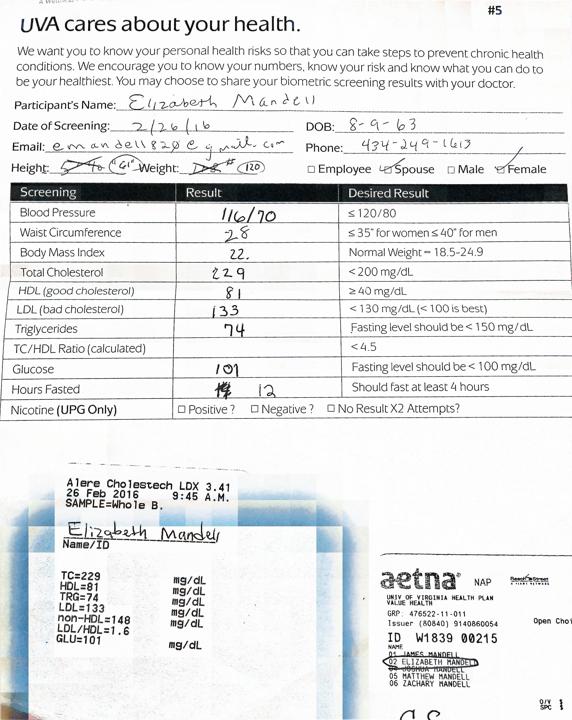}
    \caption{}
  \end{subfigure}
  \hfill
  \begin{subfigure}[b]{0.12\textwidth}
    \includegraphics[width=\textwidth]{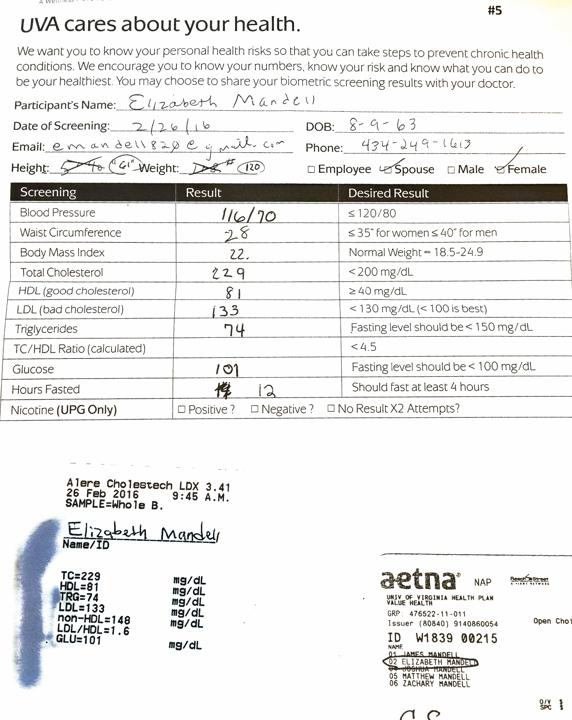}
    \caption{}
  \end{subfigure}
  \hfill
  \begin{subfigure}[b]{0.12\textwidth}
    \includegraphics[width=\textwidth]{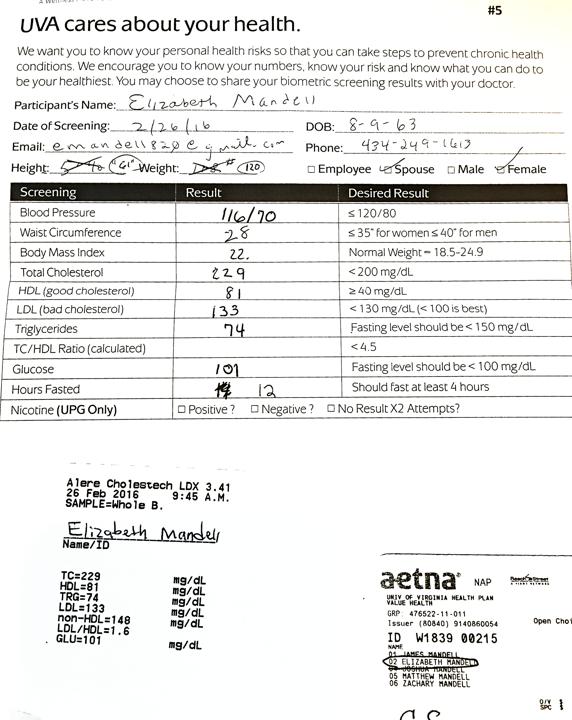}
    \caption{}
  \end{subfigure}
  
\caption{\textbf{Qualitative Comparison with State-of-the-Art DIE Methods.} (a) Original degraded images; (b) Ground-truth
reference images; (c) DocProj \cite{li2019document}; (d) DocRes \cite{Zhang_2024_CVPR}; (e) DocTr \cite{feng2021doctr}; (f) GCDRNet \cite{zhang2023appearance}; (g) Proposed GL-PGENet. Quantitative evaluation demonstrates the superior performance of our method in both structural preservation (particularly document detail enhancement and semantic legibility as shown in Row 5) and color processing (effective restoration and balanced color reproduction observed in Rows 3-4). Comparative results indicate that GL-PGENet achieves comprehensive improvements over existing benchmark methods across multiple perceptual criteria.}
  \label{fig:6x7grid}
\end{figure*}

\section{Proposed Approach}
\subsection{Overview}
We propose the GL-PGENet, a novel end-to-end deep neural network framework designed to address multiple degradation restoration in color document images.  As depicted in Fig. \ref{fig:framework}, the architecture follows a coarse-to-fine two-stage paradigm: (1) global perception parameter estimation through the Global Perception Parameter Network (GPPNet), followed by (2) local feature refinement via the Dual-Branch Local-Refine Network(DB-LRNet). Initially, a parameter regression model is employed to estimate enhancement coefficients for brightness, contrast, and saturation transformations \cite{ke2022harmonizer}, enabling efficient image-level adaptation through parallel application of these fundamental operations. This lightweight parameterization scheme ensures rapid model convergence while producing globally enhanced image $I_g$ with improved illumination consistency. Subsequently, the DB-LRNet employs dual branch network for local feature enhancement: one branch is aimed to smooth the image via convolutional networks, while the other learns linear transformations through a dense block-integrated NestUNet \cite{zhou2018unet++,huang2017densely}. The enhanced image $I_e$ is synthesized by fusing the dual outputs, preserving high-frequency details which is critical for document analysis while reinforcing local contextual consistency through adaptive parameter learning.

\subsection{Global Perception Parameter Network}
\label{sec:GPPNet}
Building upon the Lambertian reflectance assumption in GCDRNet \cite{zhang2023appearance}, where source images can be decomposed as $I = R \otimes S$ (with $R$ denoting reflectance maps and $S$ representing shadow maps), we observe that the original GC-Net's UNet architecture for pixel-wise shadow map estimation incurs substantial computational overhead. Drawing inspiration from parameter-efficient enhancement frameworks \cite{ke2022harmonizer}, we propose a GPPNet that predicts fundamental image operation parameters for size-agnostic processing.

Our network architecture employs three parallel processing branches for brightness, contrast, and saturation adjustments, followed by feature concatenation. This design differs from conventional cascaded approaches to ensure training stability. The implementation adopts a lightweight 15-layer convolutional backbone operating on 224×224 resolution inputs, maintaining computational efficiency while preserving enhancement quality. As demonstrated in Fig. \ref{fig:inter}, GPPNet produces visually compatible results $I_g$ with improved brightness, contrast, and saturation, improving computational efficiency compared to pixel-wise estimation methods.

\begin{figure*}[!t]
  \centering
  \begin{subfigure}[b]{0.15\textwidth}
    \includegraphics[width=\textwidth]{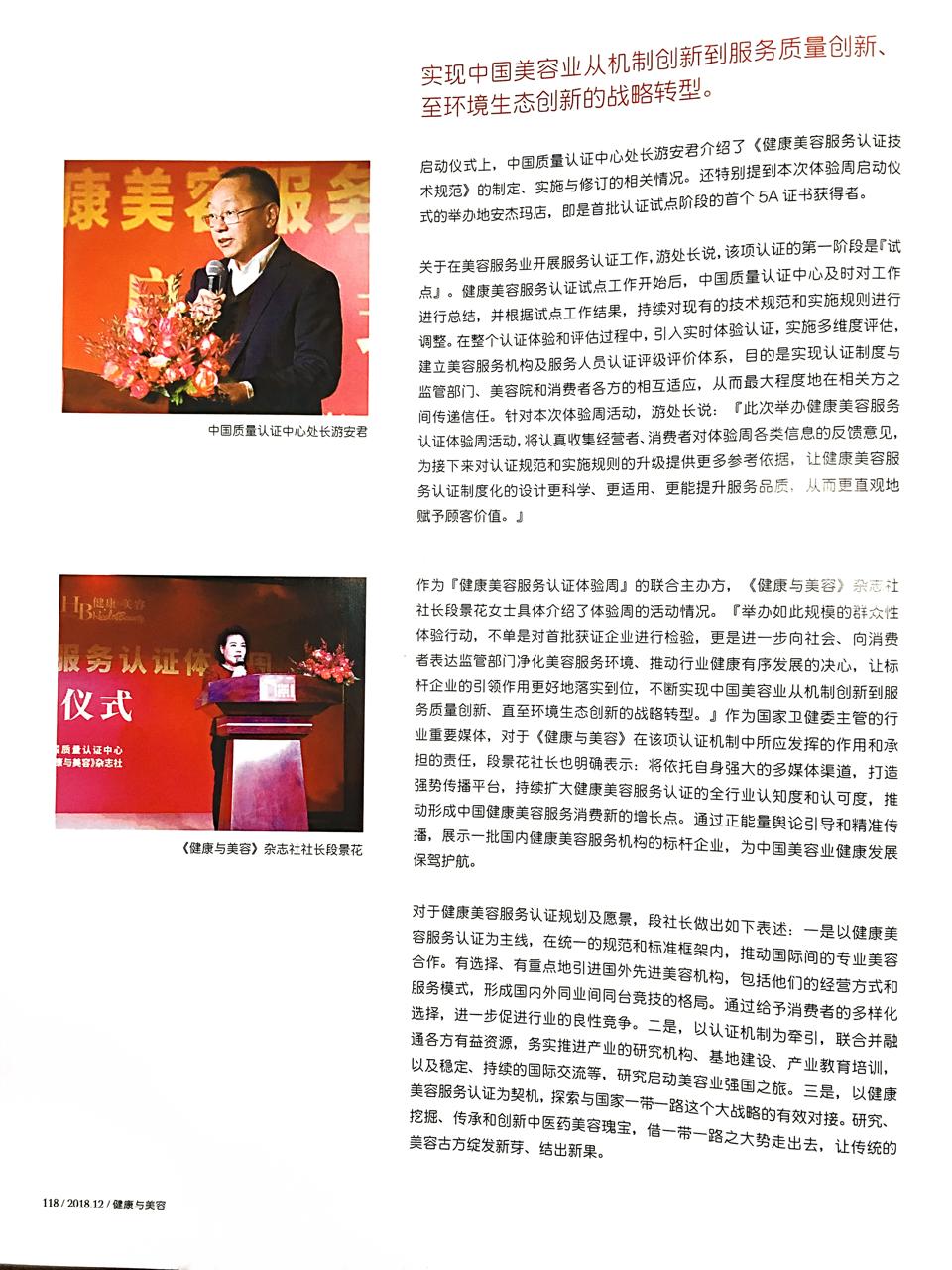}
  \end{subfigure}
  \hfill
  \begin{subfigure}[b]{0.15\textwidth}
    \includegraphics[width=\textwidth]{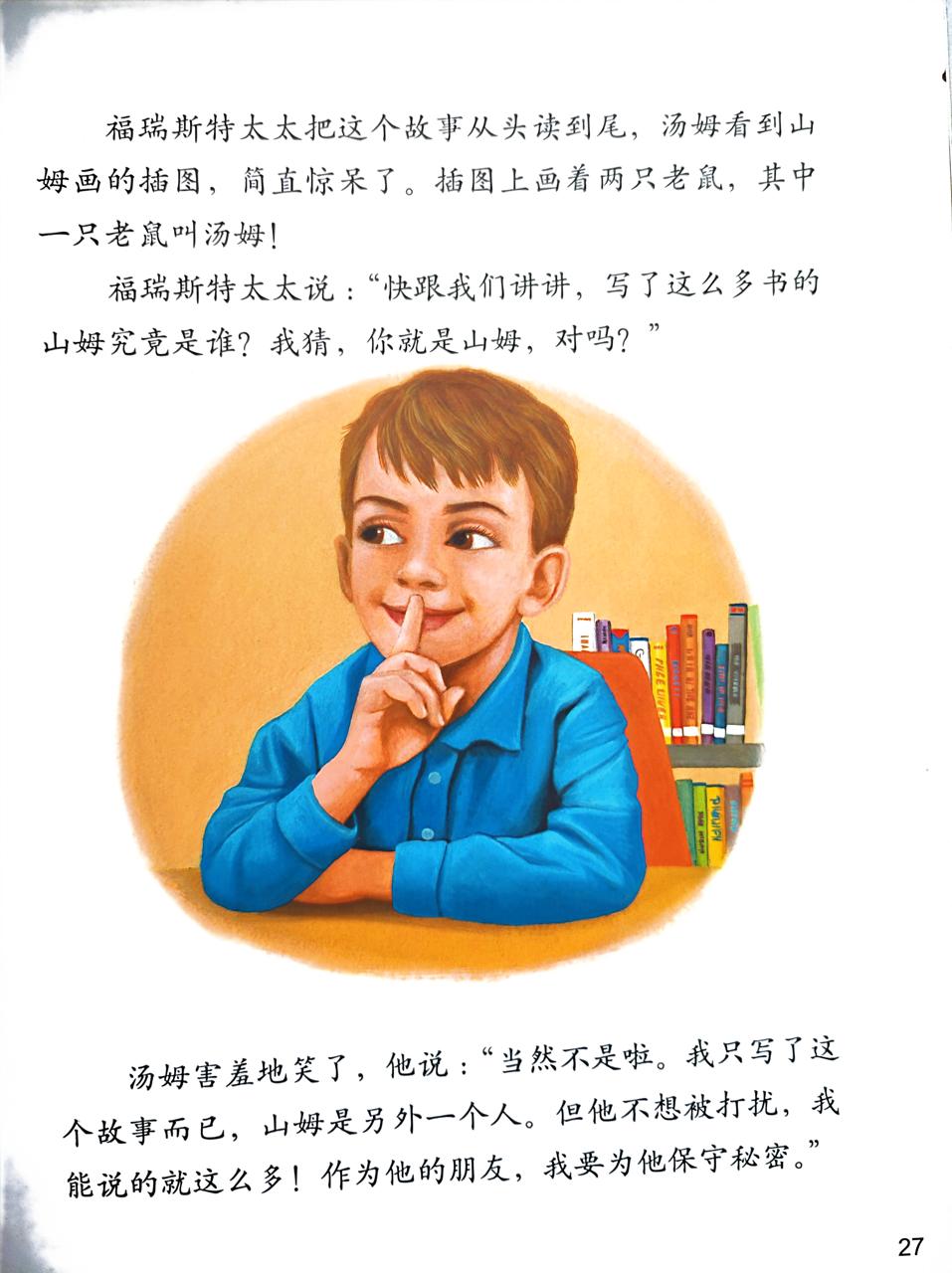}
  \end{subfigure}
  \hfill
  \begin{subfigure}[b]{0.15\textwidth}
    \includegraphics[width=\textwidth]{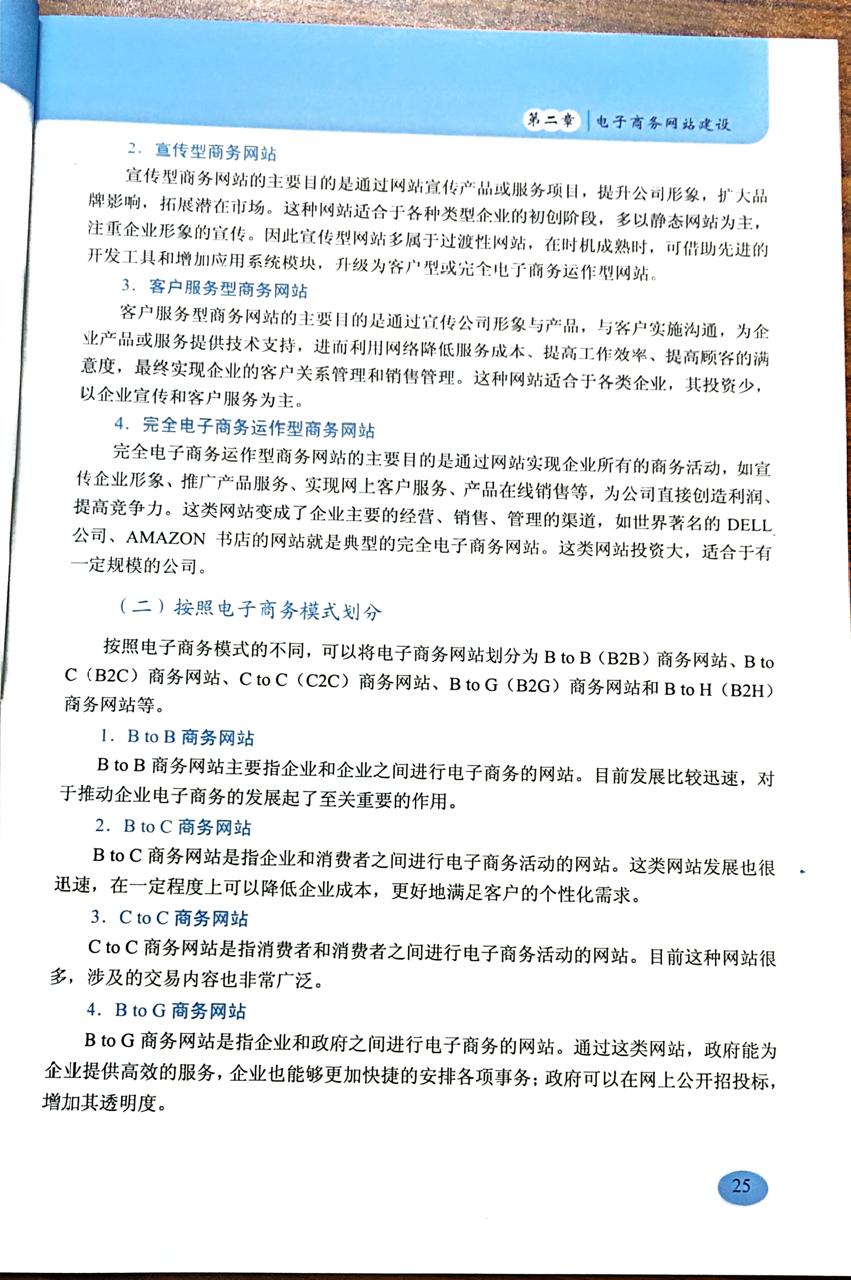}
  \end{subfigure}
  \hfill
  \begin{subfigure}[b]{0.15\textwidth}
    \includegraphics[width=\textwidth]{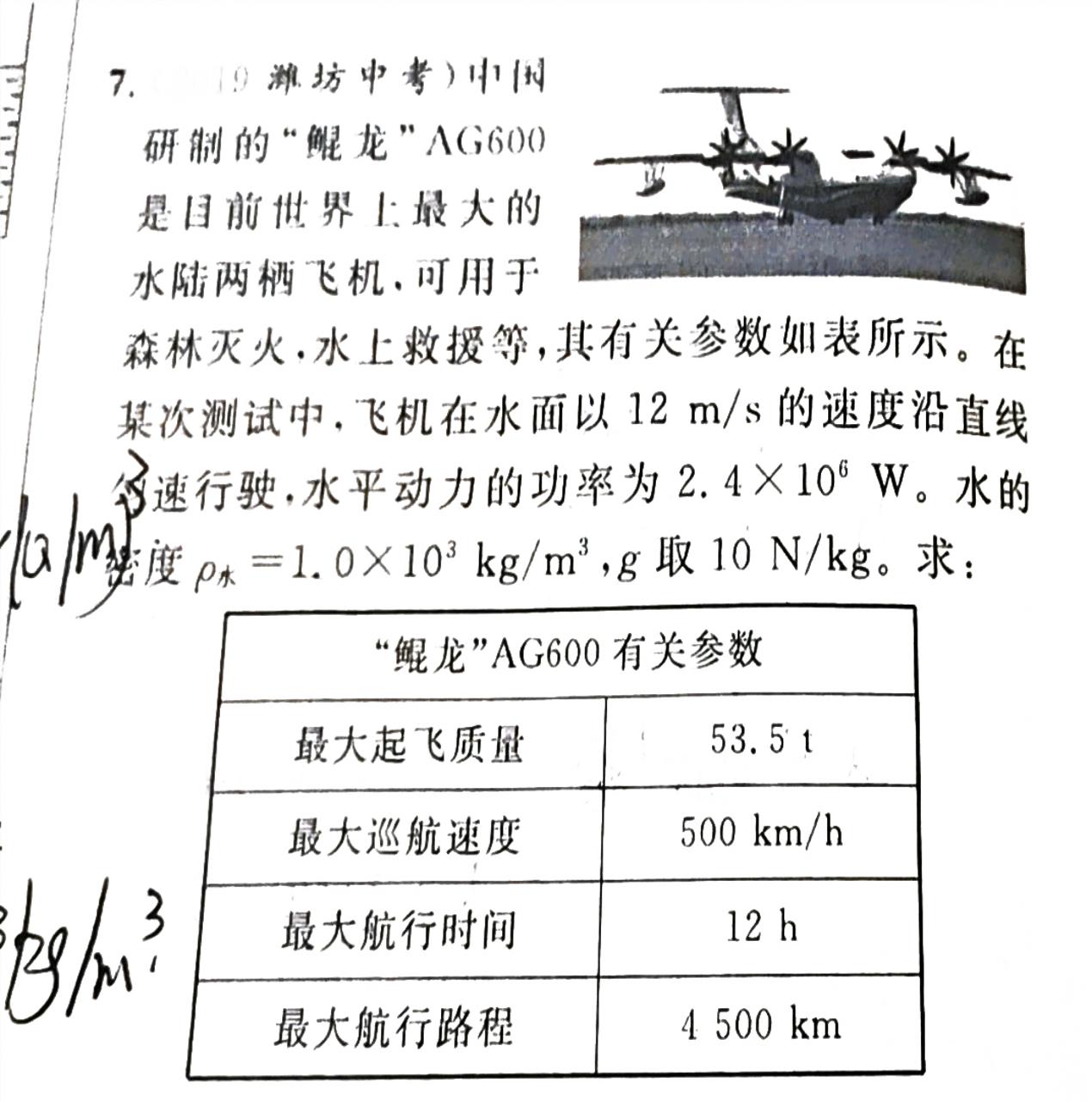}
  \end{subfigure}
  \hfill
  \begin{subfigure}[b]{0.15\textwidth}
    \includegraphics[width=\textwidth]{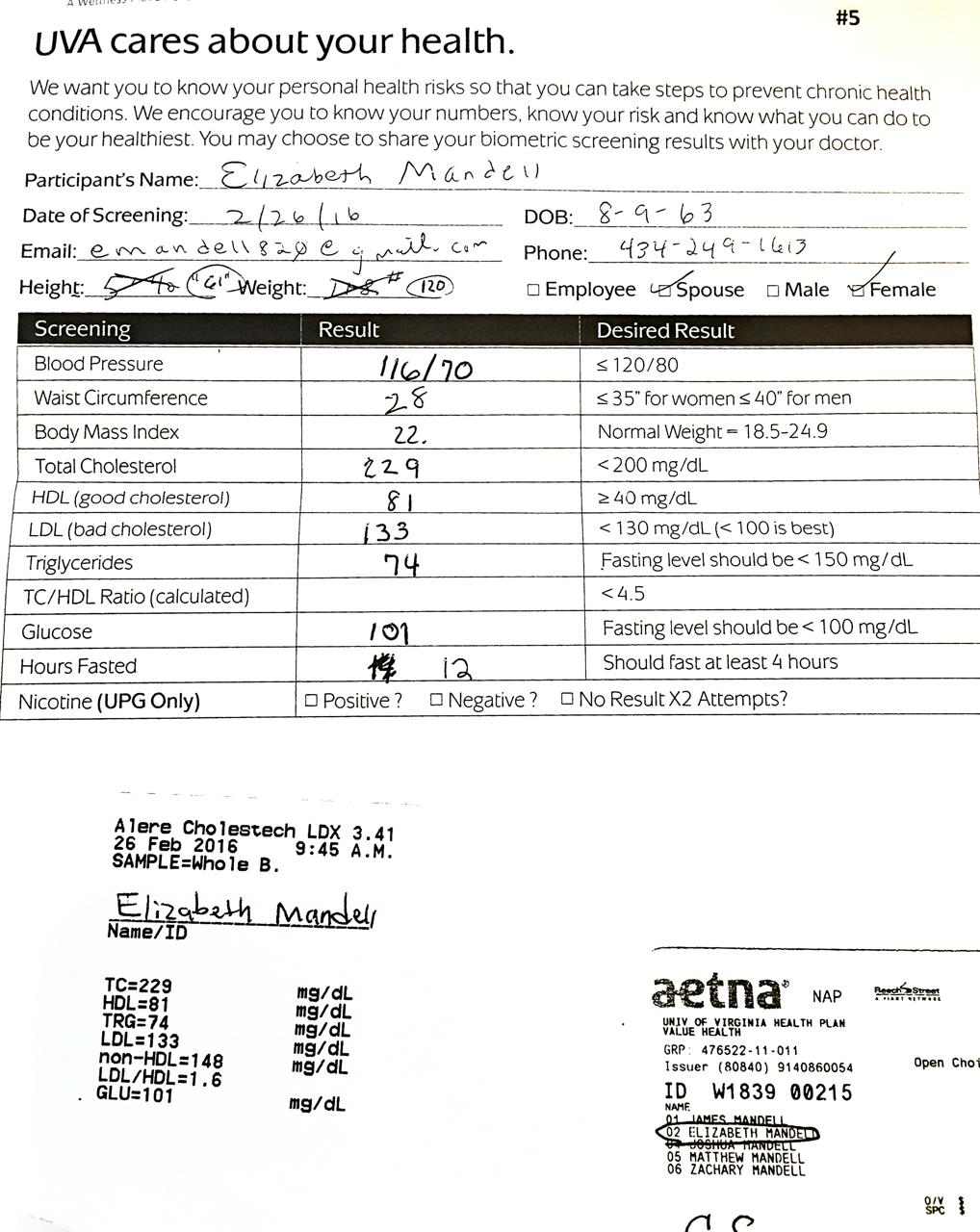}
  \end{subfigure}
  \hfill
  \begin{subfigure}[b]{0.15\textwidth}
    \includegraphics[width=\textwidth]{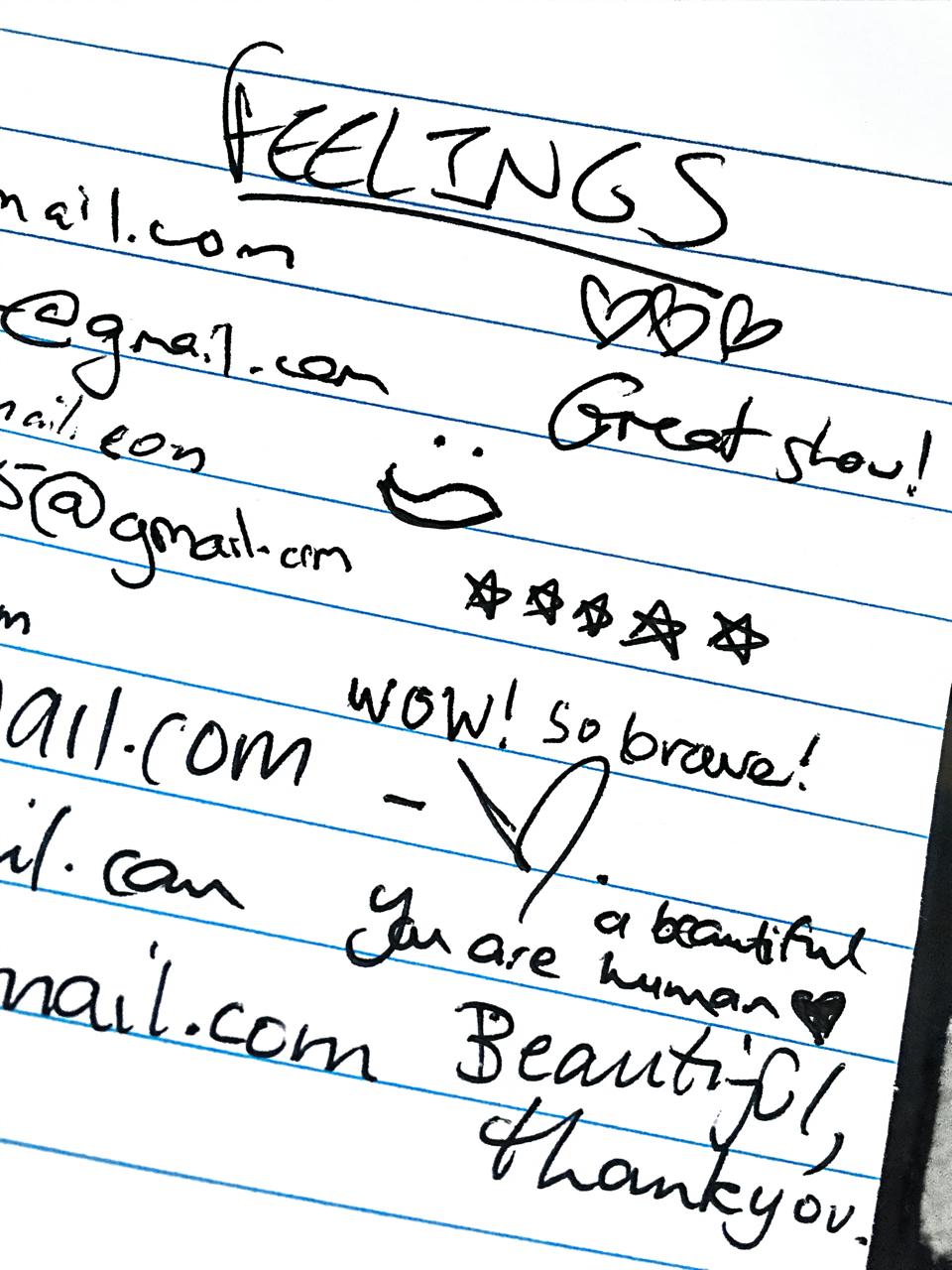}
  \end{subfigure}
  
  \begin{subfigure}[b]{0.15\textwidth}
    \includegraphics[width=\textwidth]{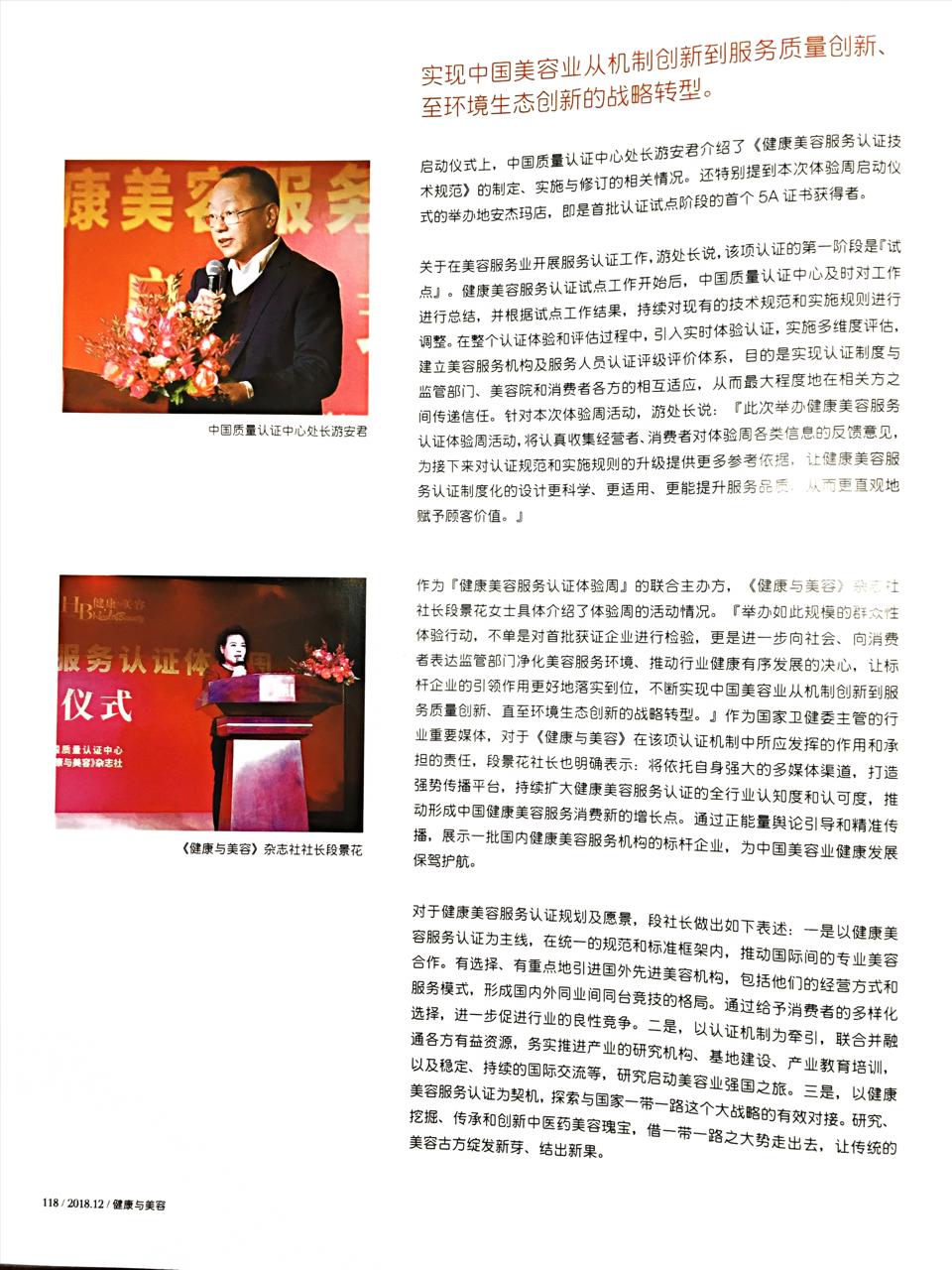}
  \end{subfigure}
  \hfill
  \begin{subfigure}[b]{0.15\textwidth}
    \includegraphics[width=\textwidth]{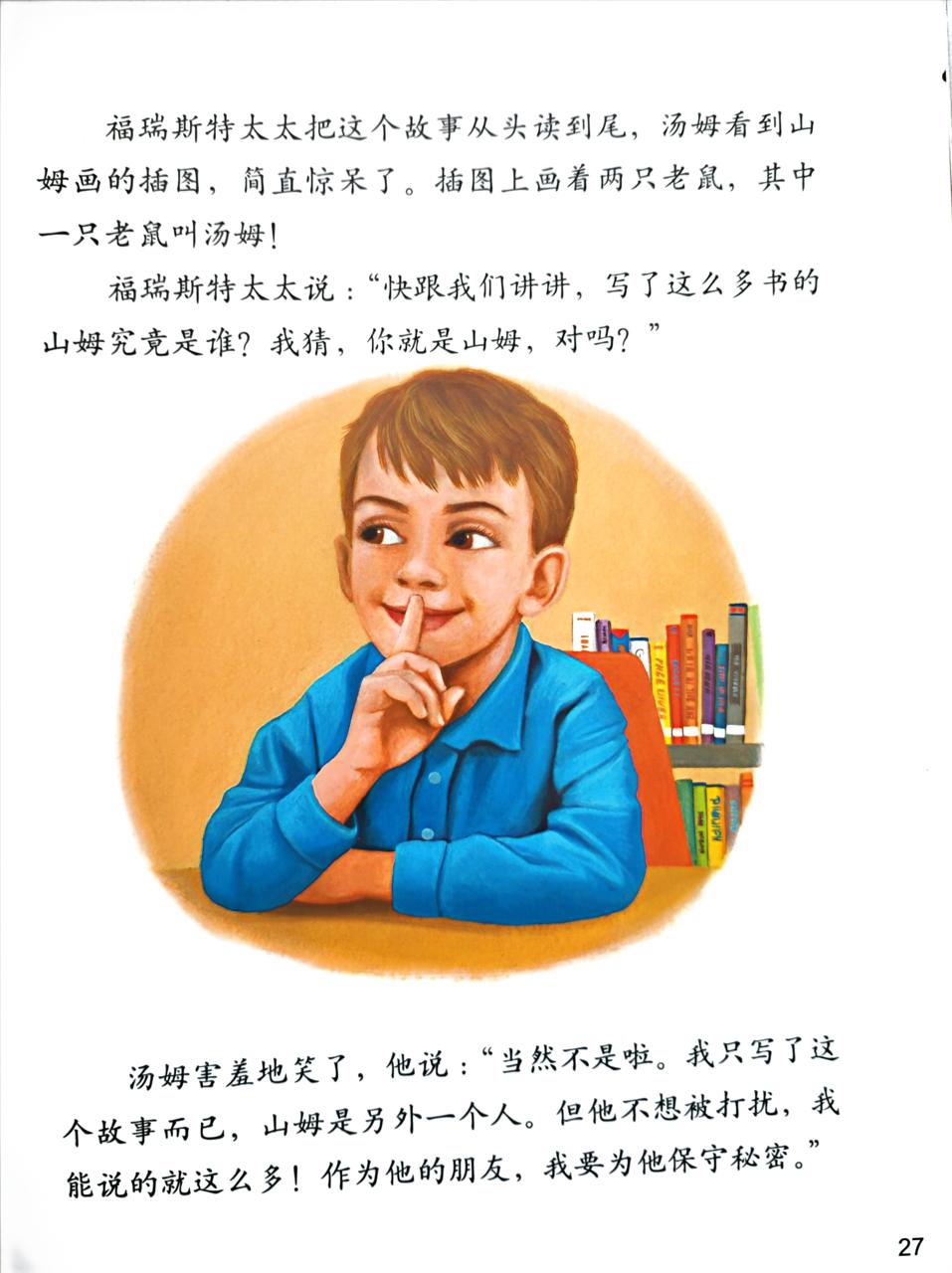}
  \end{subfigure}
  \hfill
  \begin{subfigure}[b]{0.15\textwidth}
    \includegraphics[width=\textwidth]{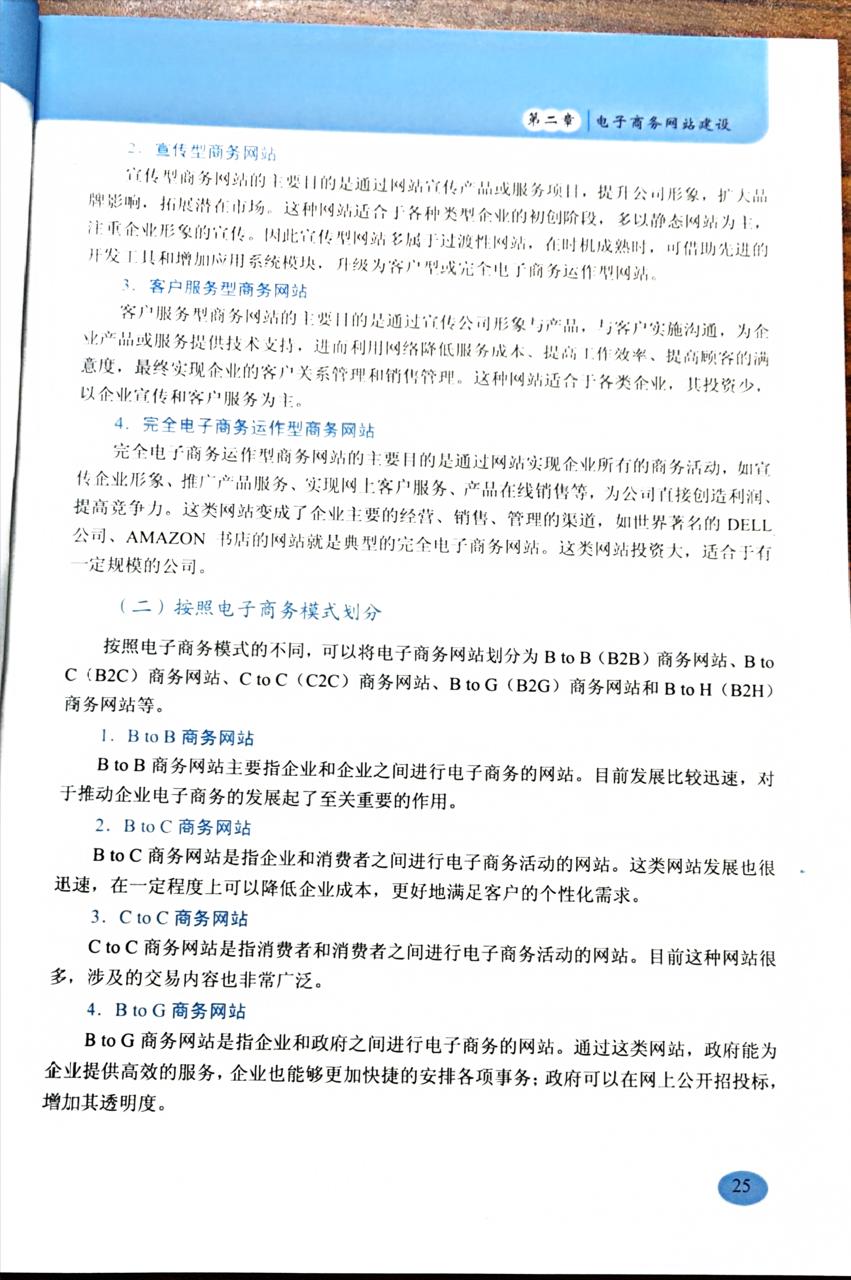}
  \end{subfigure}
  \hfill
  \begin{subfigure}[b]{0.15\textwidth}
    \includegraphics[width=\textwidth]{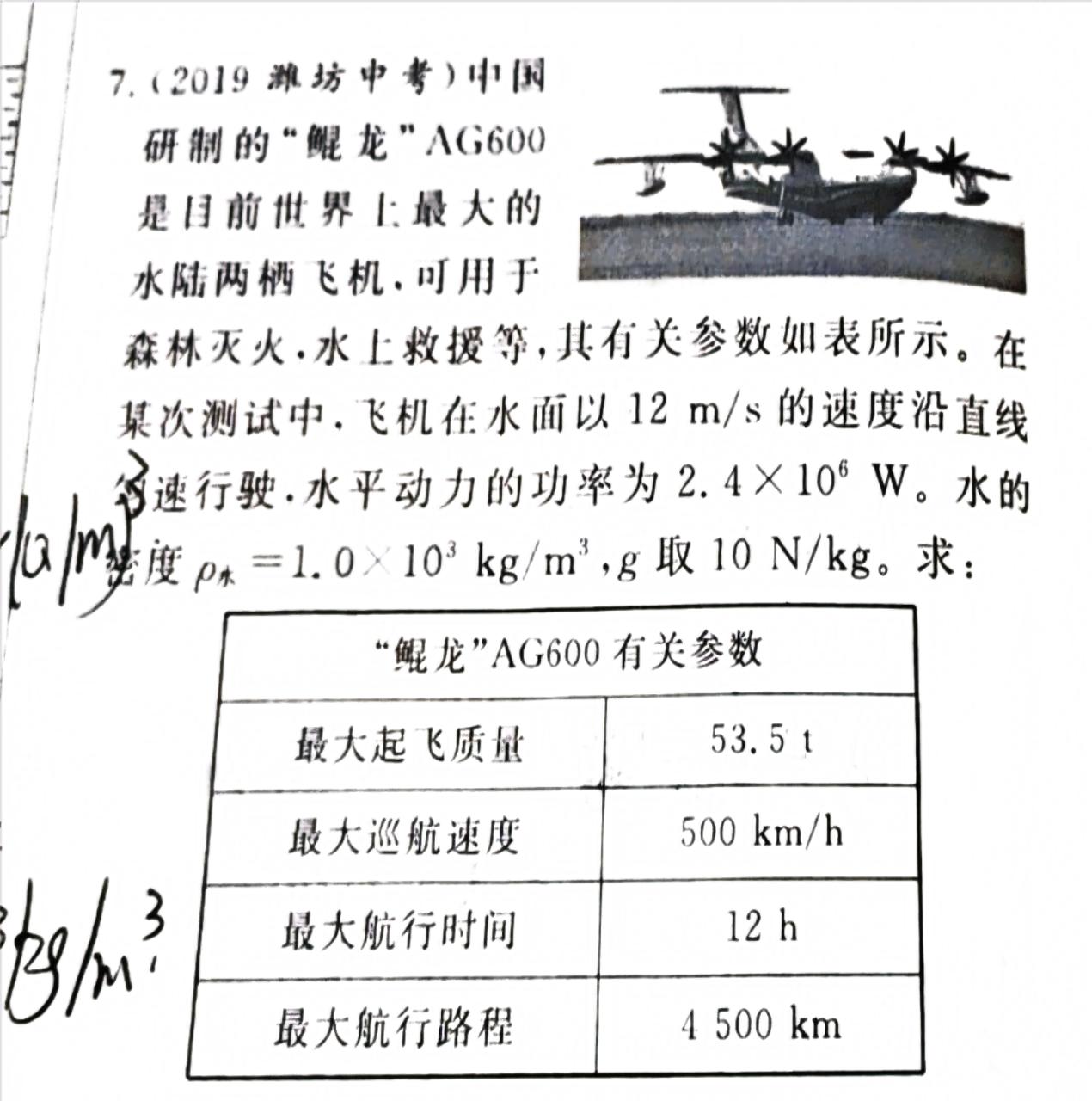}
  \end{subfigure}
  \hfill
  \begin{subfigure}[b]{0.15\textwidth}
    \includegraphics[width=\textwidth]{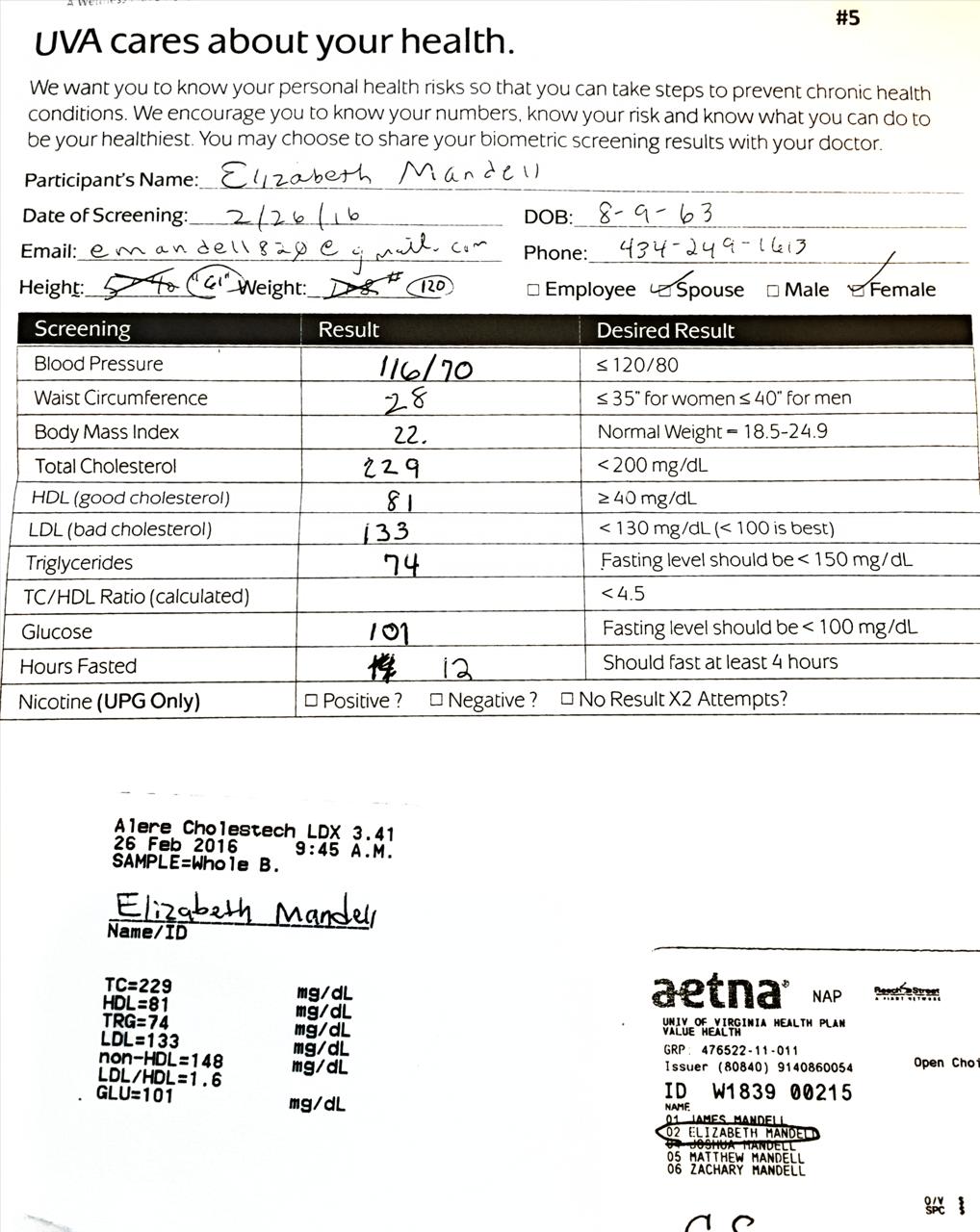}
  \end{subfigure}
  \hfill
  \begin{subfigure}[b]{0.15\textwidth}
    \includegraphics[width=\textwidth]{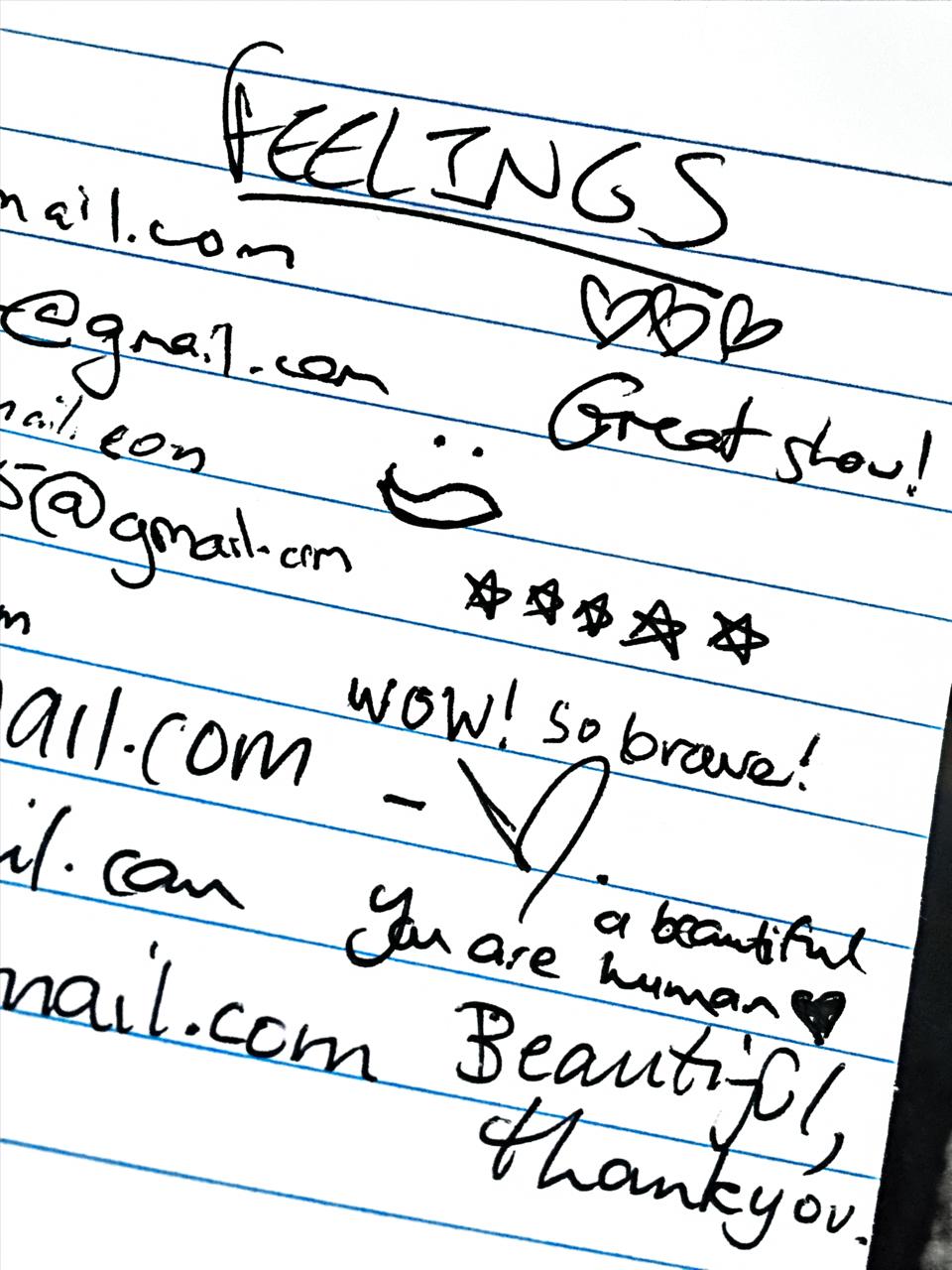}
  \end{subfigure}
  
  \caption{\textbf{Visual Comparison: Baseline vs. Our Efficient Method}. Visual comparison of inference approaches with baseline method (top row) versus our efficient implementation (bottom row), demonstrating negligible quality degradation despite $3\times$ acceleration. Quantitative analysis reveals only 1.06\% relative SSIM reduction (0.9480 vs. 0.9379) while maintaining superior performance over prior arts.}
  \label{fig:efficient_infer}
\end{figure*}

\subsection{Dual-Branch Local-Refine Network}
\label{sec:DB-LRNet}
Document image enhancement presents dual challenges: preserving fine-grained features while ensuring semantic awareness. DB-LRNet addresses this through a modified Nest UNet \cite{zhou2018unet++} integrated with dense block \cite{huang2017densely}, specifically designed to preserve critical edge features and texture patterns during semantic processing. To optimize computational efficiency, we introduce a pixel-unshuffle \cite{shi2016real} operation at the input stage, achieving 2$\times$ spatial reduction through learned sub-pixel reorganization while maintaining information completeness. This preprocessing strategy preserves original content fidelity with significantly reduced computational complexity.

As illustrated in Fig. \ref{fig:framework}, our framework introduces a novel dual-branch paradigm that diverges from conventional generative approaches where enhanced images $I_e$ are directly synthesized through network $f(\cdot)$ as in Eq. \ref{eq:eq1}. The first branch employs three convolutional layers for smoothing the image, while the second branch predicts linear transformation parameters through the modified Nest UNet \cite{zhou2018unet++} with dense block. The enhanced output is formulated through Eq. \ref{eq:eq2}-\ref{eq:eq4}, combining these two branches. This architecture provides two principal advantages: (1) enhanced local consistency achieved by constraining linear transformation parameters through Eq. \ref{eq:eq6}, which generates smoother output through parameter regularization, and (2) improved computational efficiency for high-resolution processing with negligible quality degradation. These technical advancements, including implementation details and comprehensive evaluations, will be elaborated in Sec. \ref{sec: efficient_inference}.
\begin{equation}
I_L = f(I_O)
\label{eq:eq1}
\end{equation}

\begin{equation}
feat = f(I_O)
\label{eq:eq2}
\end{equation}

\begin{equation}
\alpha = g_1(feat), \quad \beta = g_2(feat)
\label{eq:eq3}
\end{equation}

\begin{equation}
I_L = \alpha h(I_O) + \beta
\label{eq:eq4}
\end{equation}

\section{Experiments}
\subsection{Implementation details}
The proposed framework is implemented in PyTorch \cite{paszke2017automatic} and trained on a single NVIDIA A100-40G GPU with a batch size of 16. To ensure training stability, we adopt a two-stage optimization strategy: (a) GPPNet undergoes independent pre-training to establish reliable $I_g$ estimation before initiating joint optimization with the DB-LRNet. (b) During joint training, gradient backpropagation to GPPNet remains disabled to preserve parameter stability. The Adam optimizer with initial learning rate 0.0001($\beta_1=0.9$, $\beta_2=0.99$) drives the optimization process. The architecture accepts dual-resolution inputs: GPPNet processes $224 \times 224$ images while DB-LRNet operates at $512\times 512$ resolution.

During pre-training, both networks are trained on a dataset of 500,000+ synthetic samples by minimizing a composite loss function which integrates adversarial loss \cite{goodfellow2020generative}, L1 loss, structural similarity (SSIM) \cite{wang2004image} and total variation (TV) loss \cite{aly2005image}:
\begin{equation}
L = \lambda_1 L_1 + \lambda_2 L_{SSIM} + \lambda_3 L_{TV} + \lambda_4 L_{GAN} + \lambda_5 L_r
\label{eq:eq5}
\end{equation}
Here we set $\lambda_1$, $\lambda_2$, $\lambda_3$, $\lambda_4$ and $\lambda_5$ to 1, 0.5, 0.01, 0.05 and 0.01. Notably, the adversarial loss is excluded during fine-tuning to emphasize perceptual quality preservation, while introducing an additional regularization term: 
\begin{equation}
L_r = \lVert \nabla \alpha \rVert^2 + \lVert \nabla \beta \rVert^2
\label{eq:eq6}
\end{equation}
This multi-stage training paradigm significantly enhances model performance on document image enhancement tasks. All reported results adhere to the standardized evaluation protocol using the RealDAE test set unless otherwise specified.

\begin{figure}[t]
    \centering
    \includegraphics[width=0.45\textwidth]{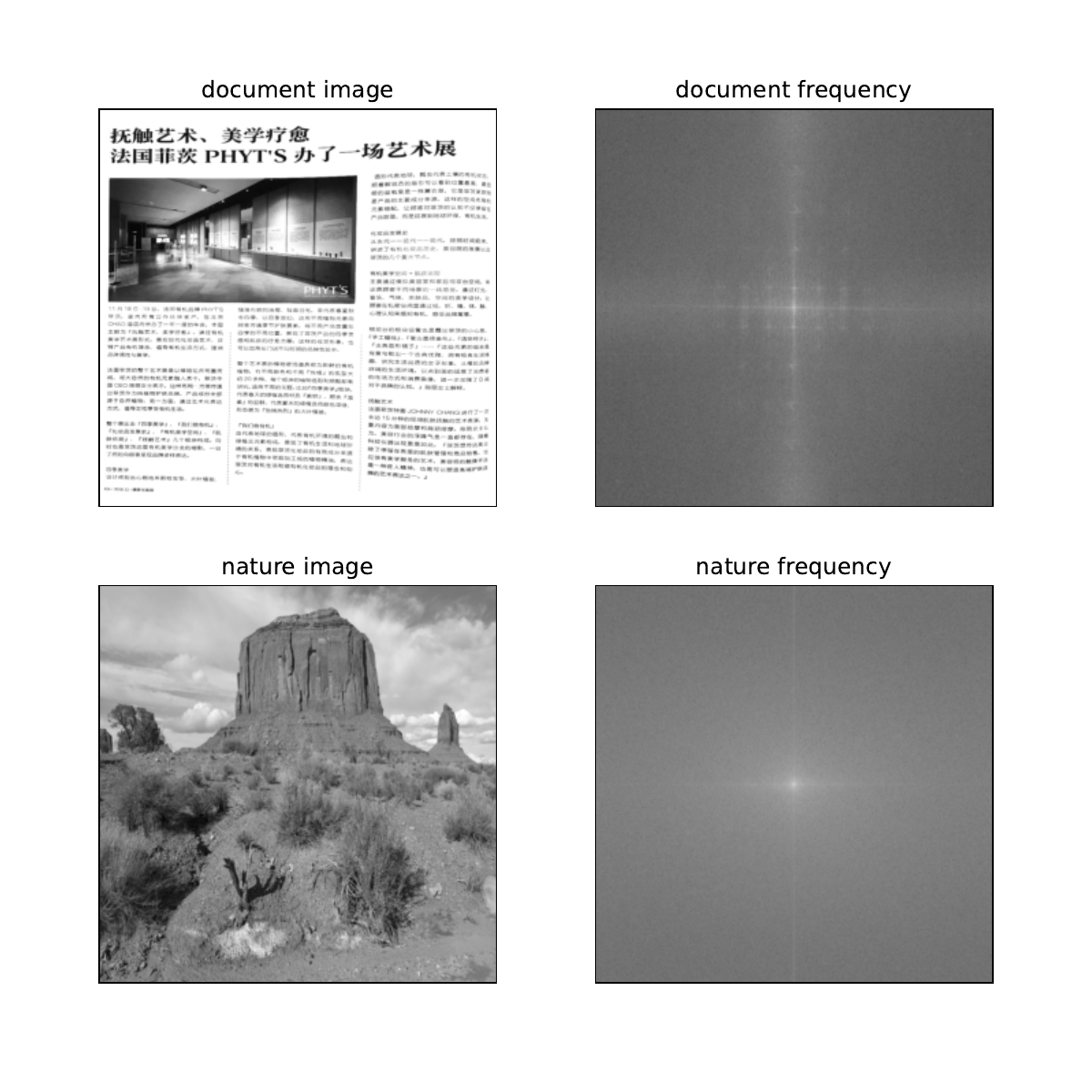}
    \caption{\textbf{Frequency Analysis of Natural Images and Document Images.} There are obvious high-frequency components in the horizontal and vertical directions in the document image, but the energy in nature image is mainly concentrated in the low frequency part.}
    \label{fig:frequency_analysis}
\end{figure}

\begin{figure}[ht]
    \centering
    \includegraphics[width=0.48\textwidth]{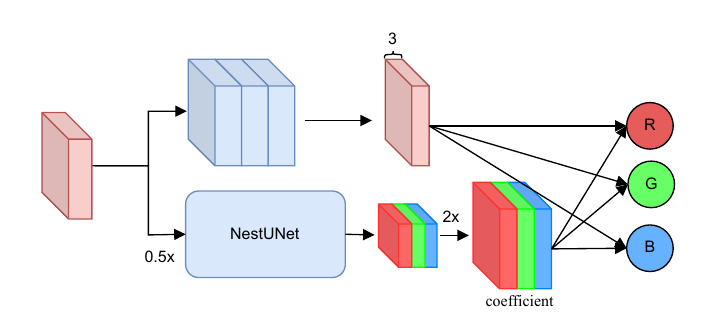}
    \caption{\textbf{Architecture of DB-LRNet's efficient inference framework.} The parameter prediction branch employs resolution reduction followed by coefficient map upsampling, achieving 75\% computational complexity reduction while maintaining performance parity. This optimization requires no additional training.}
    \label{fig:framework_lrnet}
\end{figure}

\begin{figure}[!ht]
  \centering
  \begin{subfigure}[b]{0.11\textwidth}
    \includegraphics[width=\textwidth]{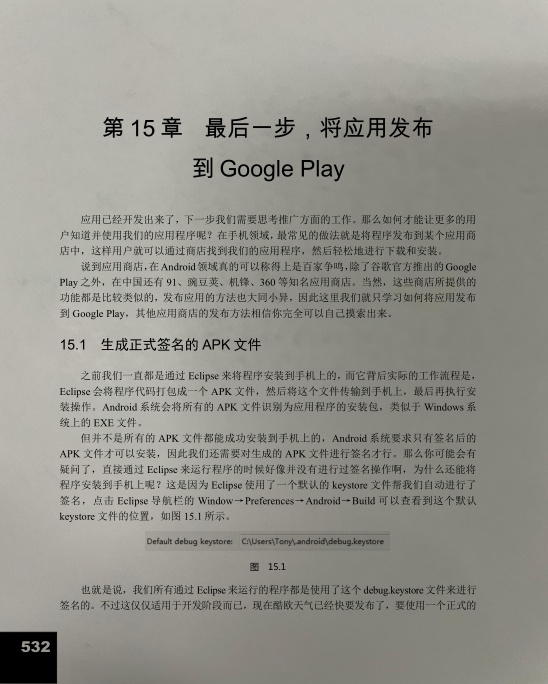}
  \end{subfigure}
  \hfill
  \begin{subfigure}[b]{0.11\textwidth}
    \includegraphics[width=\textwidth]{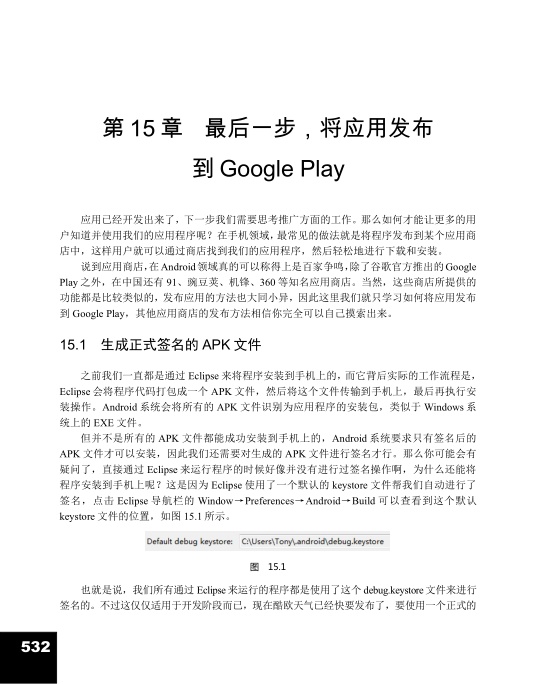}
  \end{subfigure}
  \hfill
  \begin{subfigure}[b]{0.11\textwidth}
    \includegraphics[width=\textwidth]{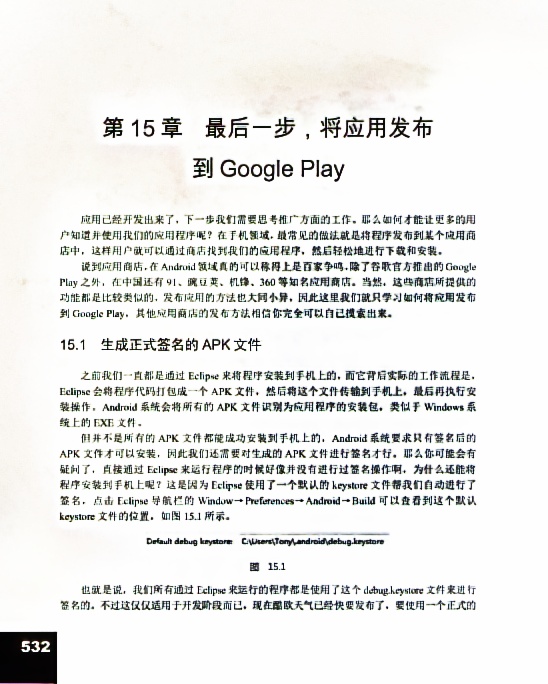}
  \end{subfigure}
  \hfill
  \begin{subfigure}[b]{0.11\textwidth}
    \includegraphics[width=\textwidth]{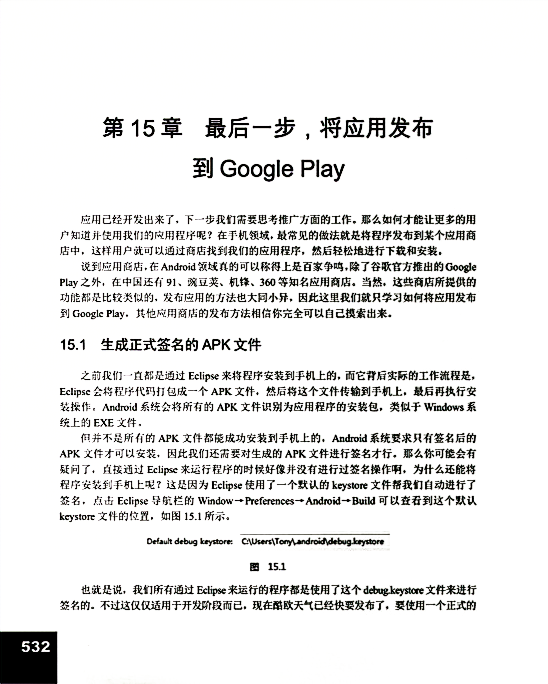}
  \end{subfigure}
  
  \begin{subfigure}[b]{0.11\textwidth}
    \includegraphics[width=\textwidth]{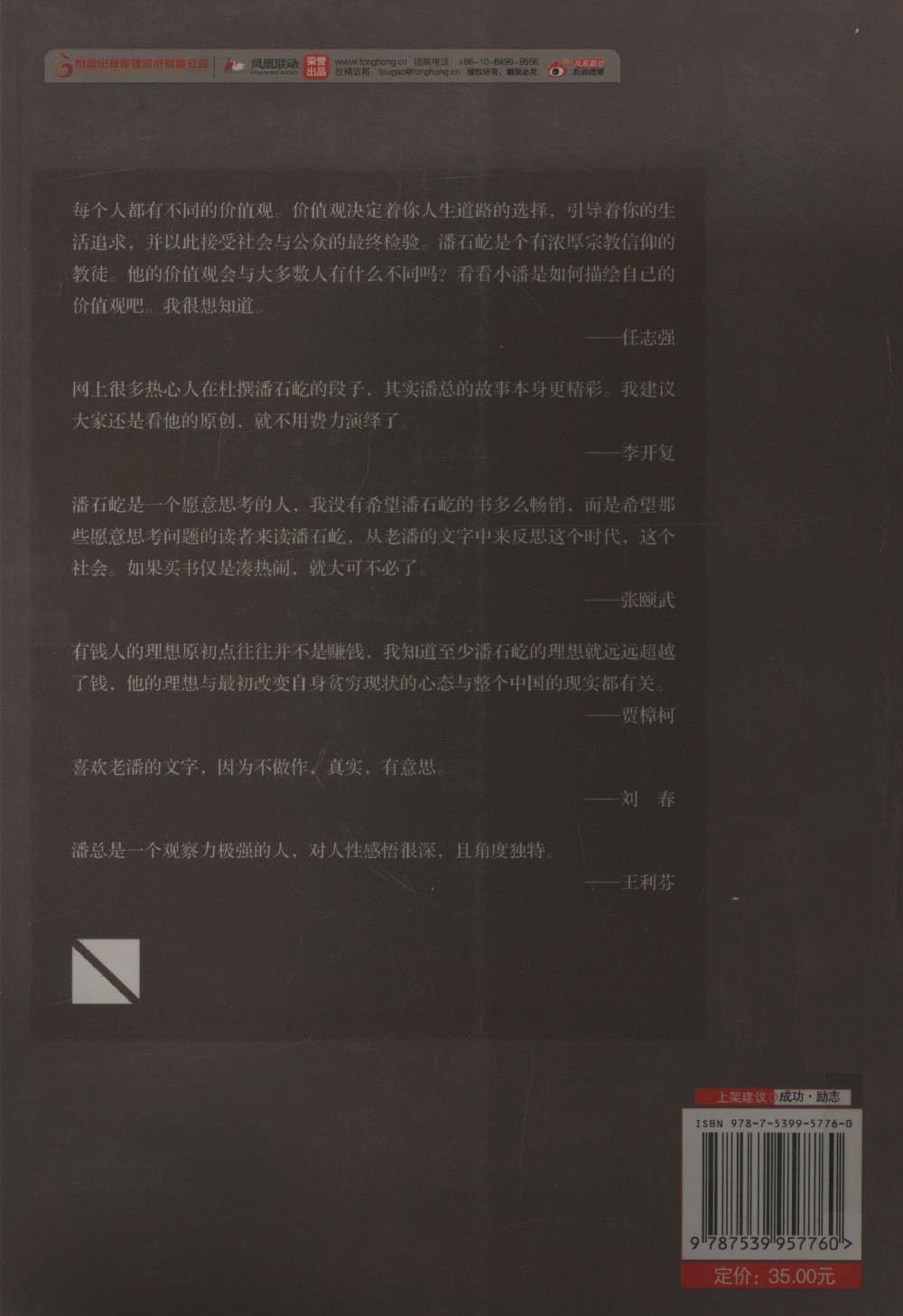}
  \end{subfigure}
  \hfill
  \begin{subfigure}[b]{0.11\textwidth}
    \includegraphics[width=\textwidth]{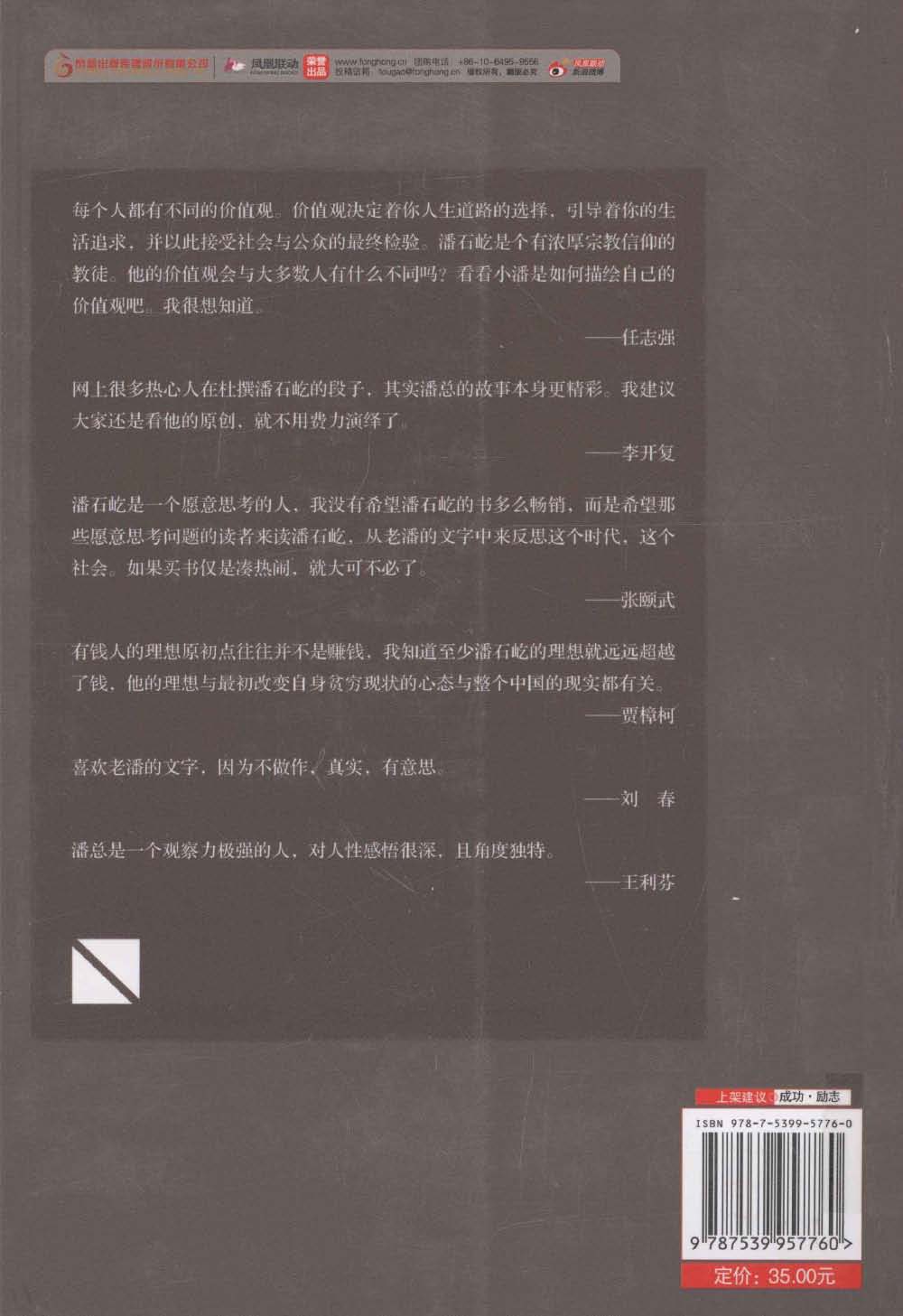}
  \end{subfigure}
  \hfill
  \begin{subfigure}[b]{0.11\textwidth}
    \includegraphics[width=\textwidth]{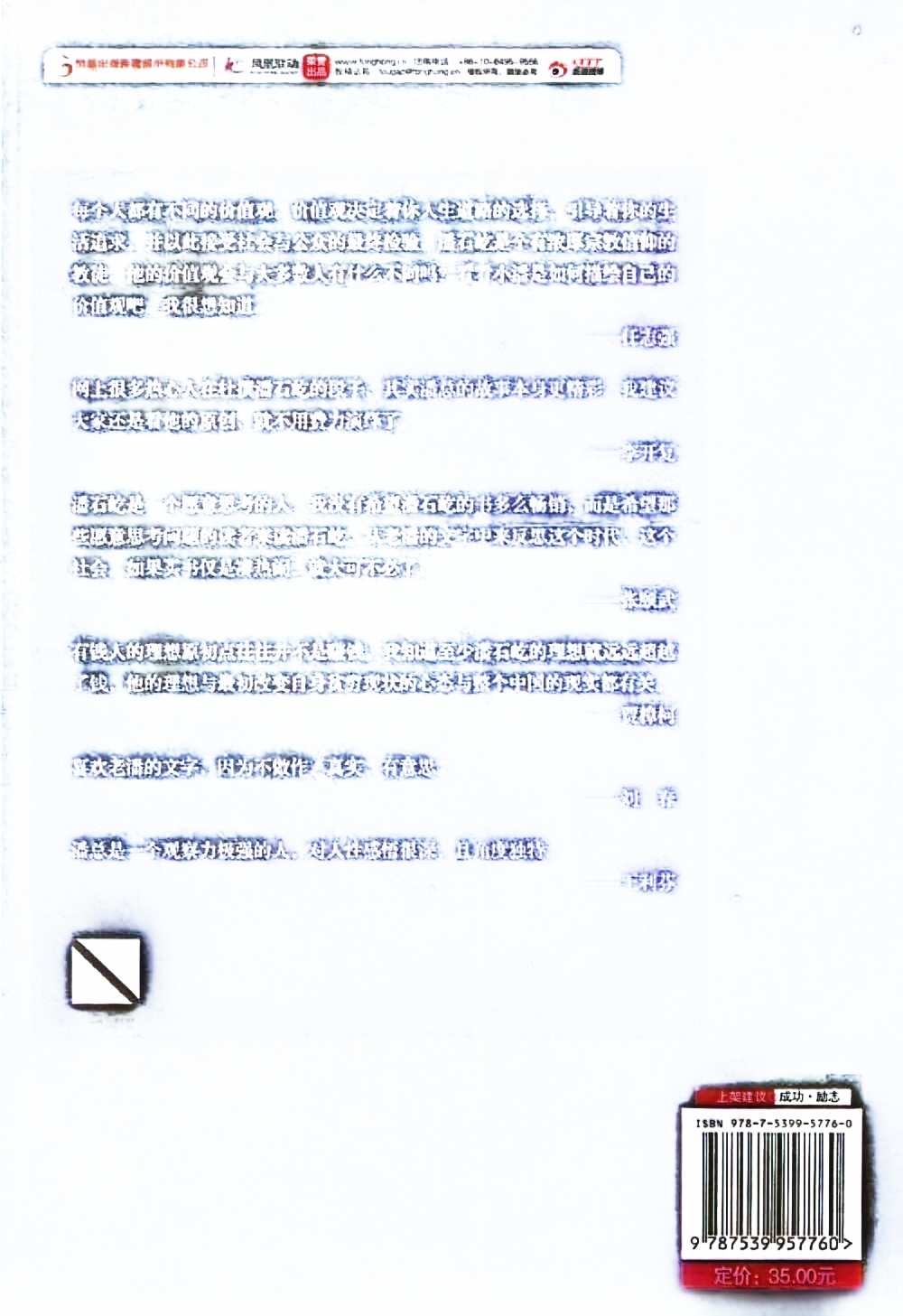}
  \end{subfigure}
  \hfill
  \begin{subfigure}[b]{0.11\textwidth}
    \includegraphics[width=\textwidth]{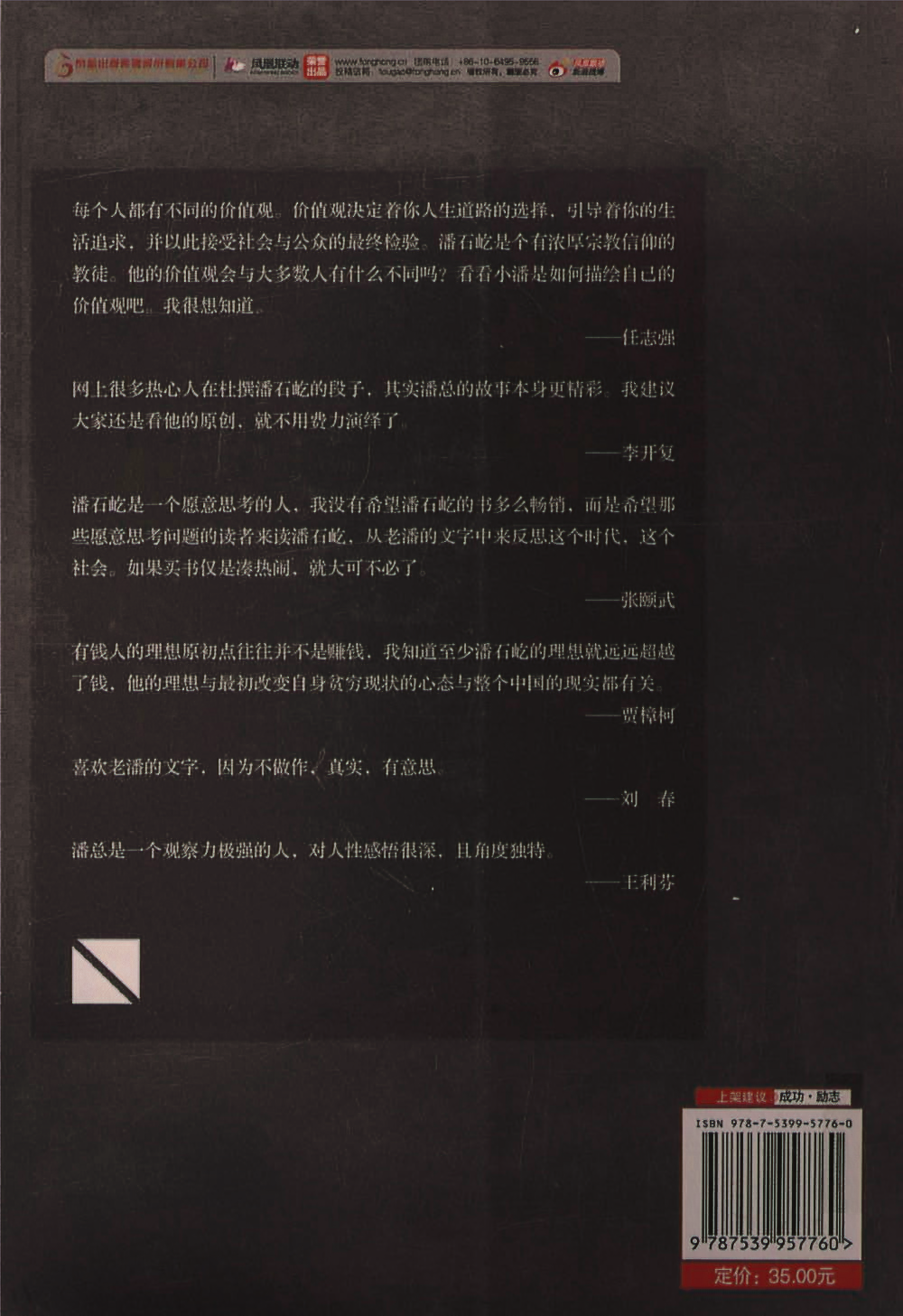}
  \end{subfigure}
  
  \begin{subfigure}[b]{0.11\textwidth}
    \includegraphics[width=\textwidth]{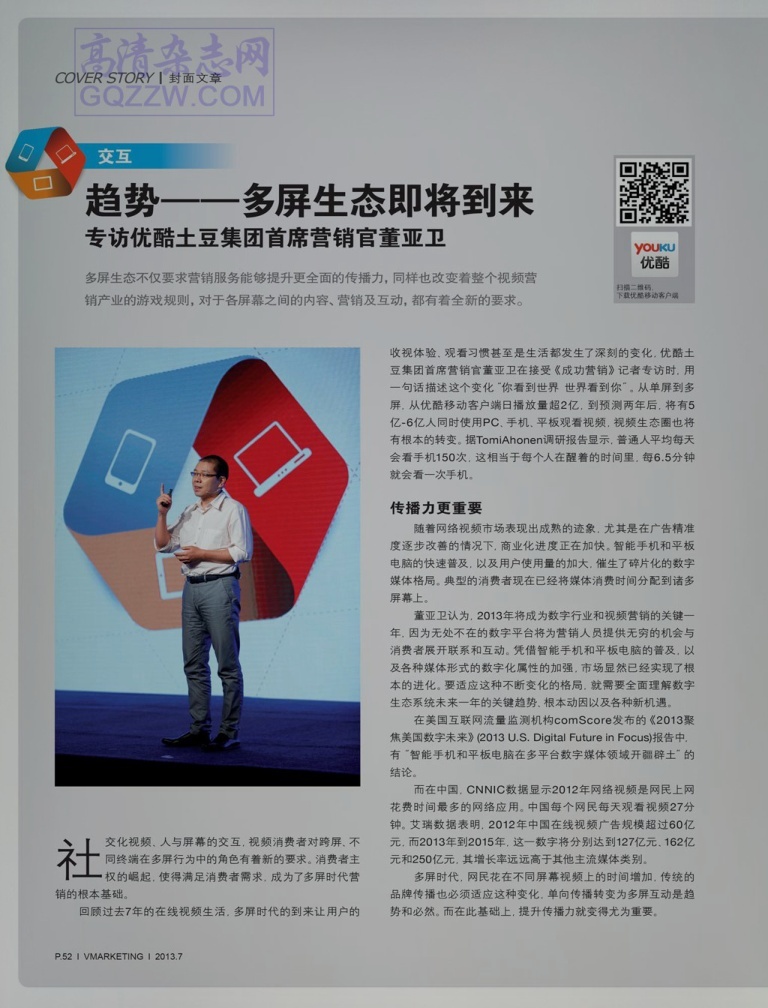}
  \end{subfigure}
  \hfill
  \begin{subfigure}[b]{0.11\textwidth}
    \includegraphics[width=\textwidth]{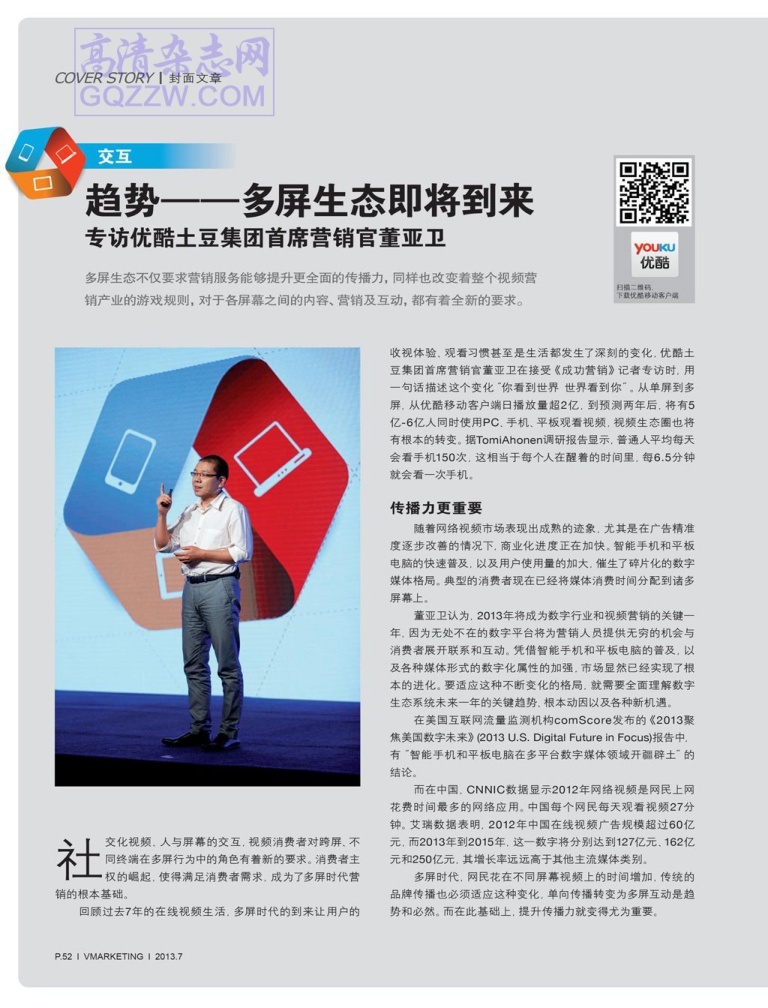}
  \end{subfigure}
  \hfill
  \begin{subfigure}[b]{0.11\textwidth}
    \includegraphics[width=\textwidth]{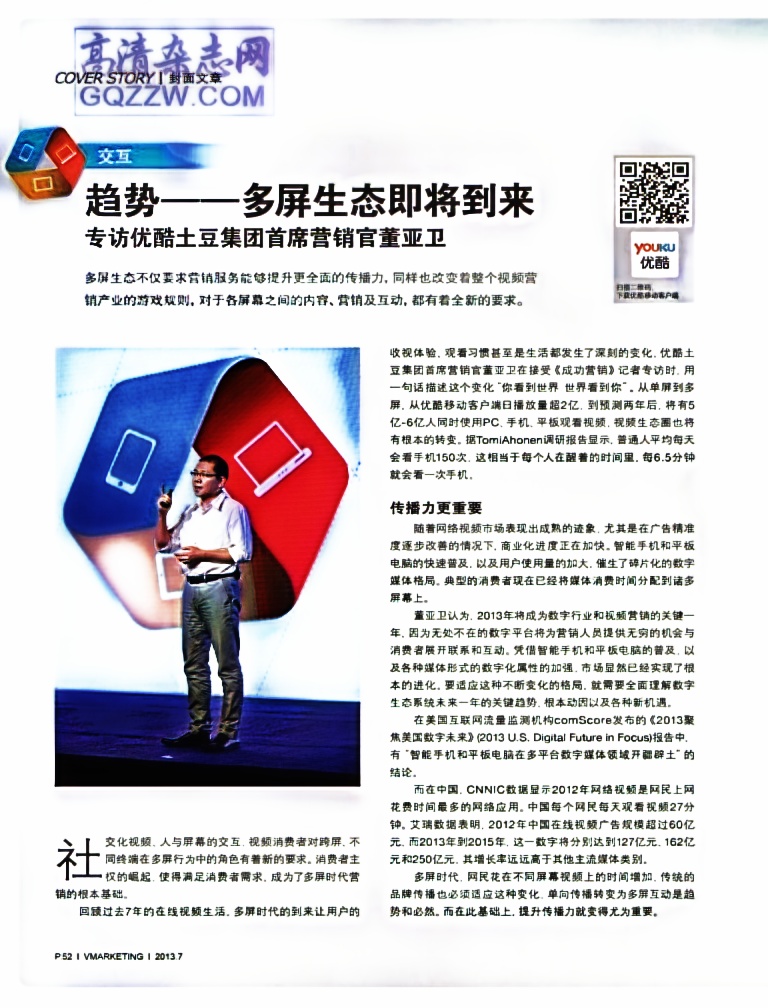}
  \end{subfigure}
  \hfill
  \begin{subfigure}[b]{0.11\textwidth}
    \includegraphics[width=\textwidth]{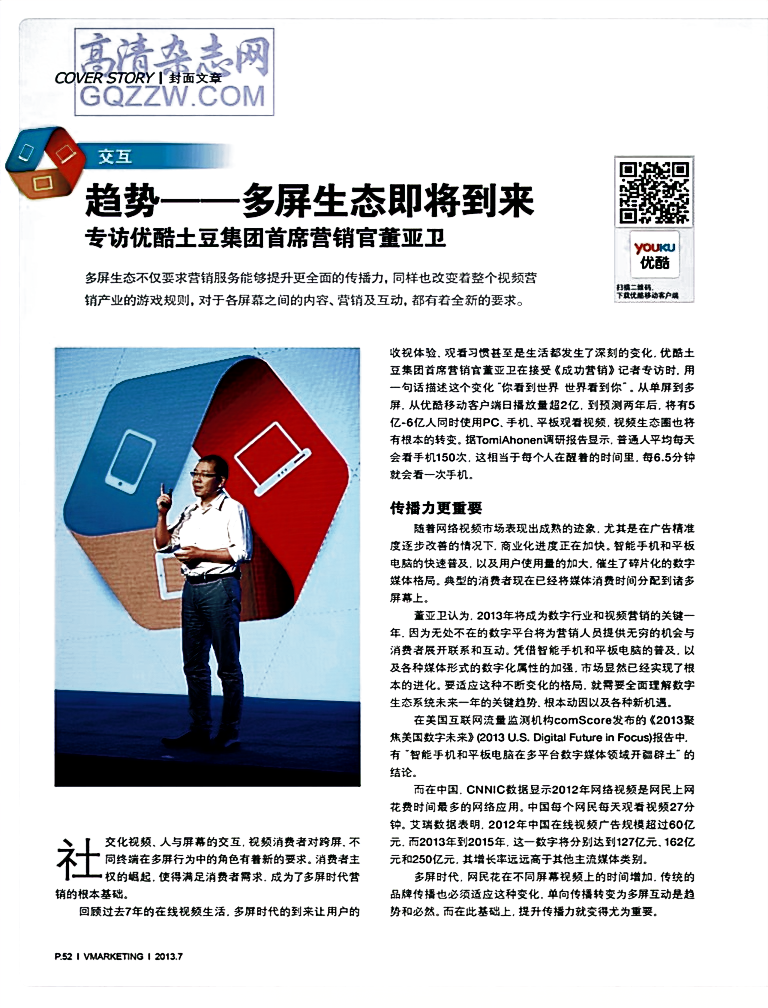}
  \end{subfigure}
  \begin{subfigure}[b]{0.11\textwidth}
    \includegraphics[width=\textwidth]{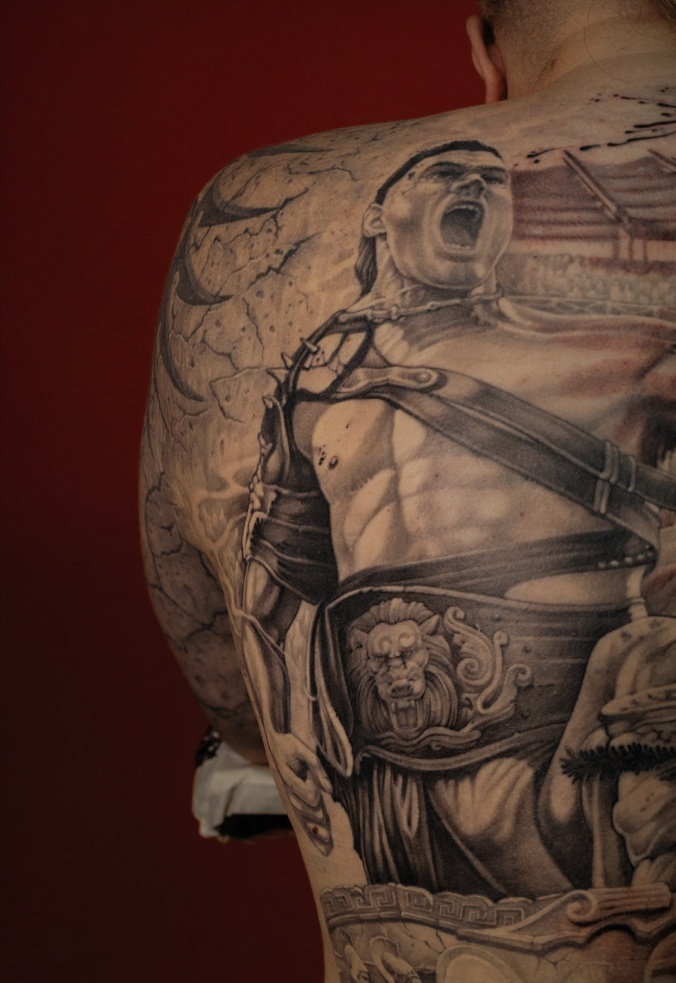}
  \end{subfigure}
  \hfill
  \begin{subfigure}[b]{0.11\textwidth}
    \includegraphics[width=\textwidth]{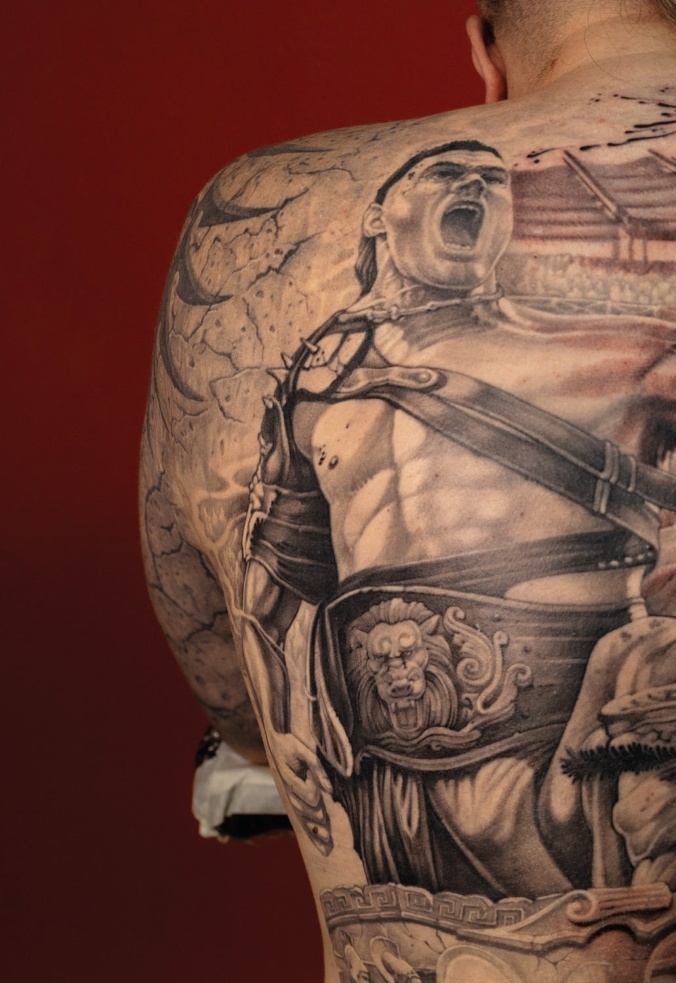}
  \end{subfigure}
  \hfill
  \begin{subfigure}[b]{0.11\textwidth}
    \includegraphics[width=\textwidth]{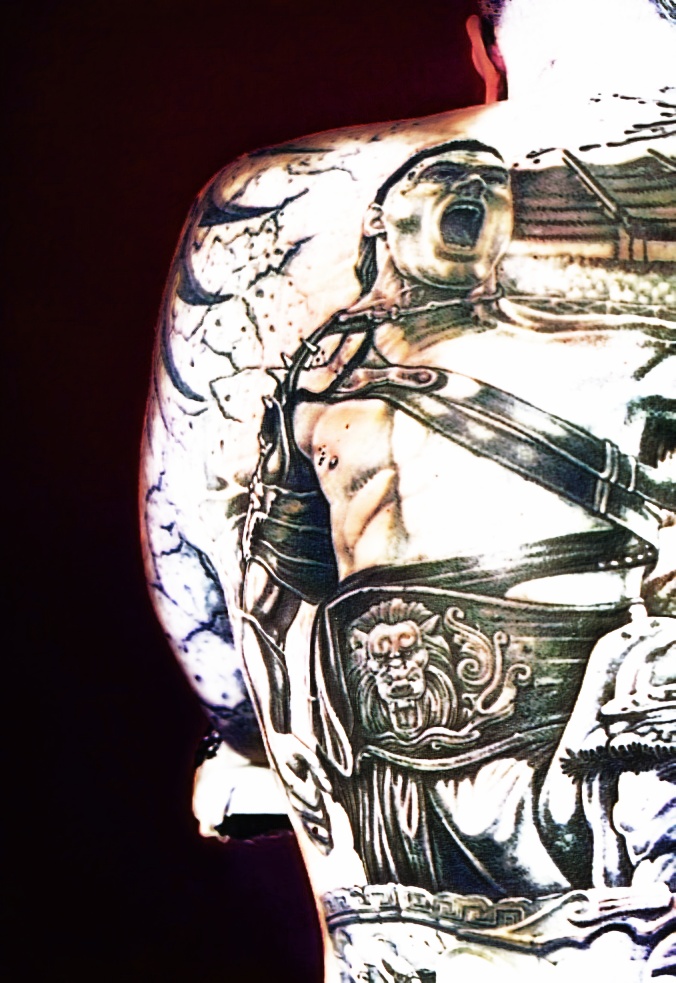}
  \end{subfigure}
  \hfill
  \begin{subfigure}[b]{0.11\textwidth}
    \includegraphics[width=\textwidth]{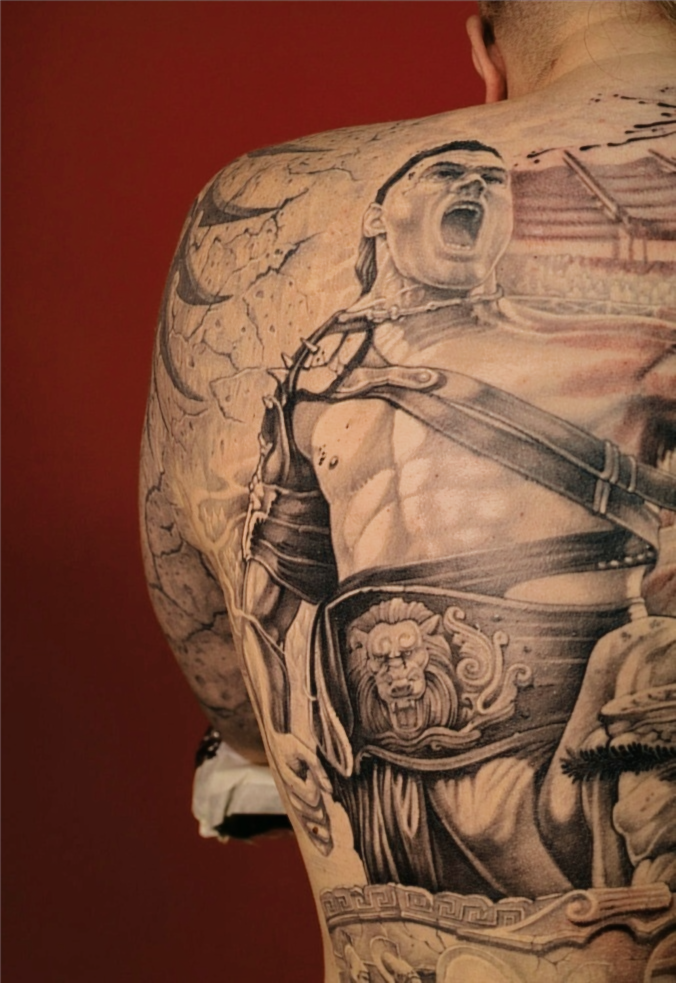}
  \end{subfigure}
  \begin{subfigure}[b]{0.11\textwidth}
    \includegraphics[width=\textwidth]{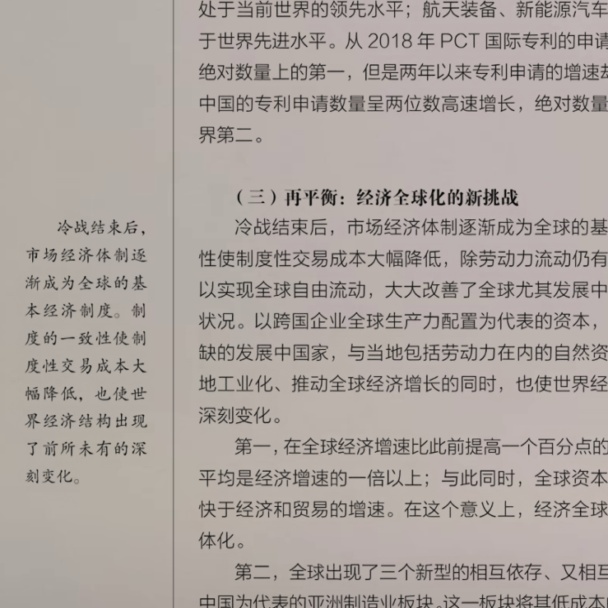}
    \caption{source}
  \end{subfigure}
  \hfill
  \begin{subfigure}[b]{0.11\textwidth}
    \includegraphics[width=\textwidth]{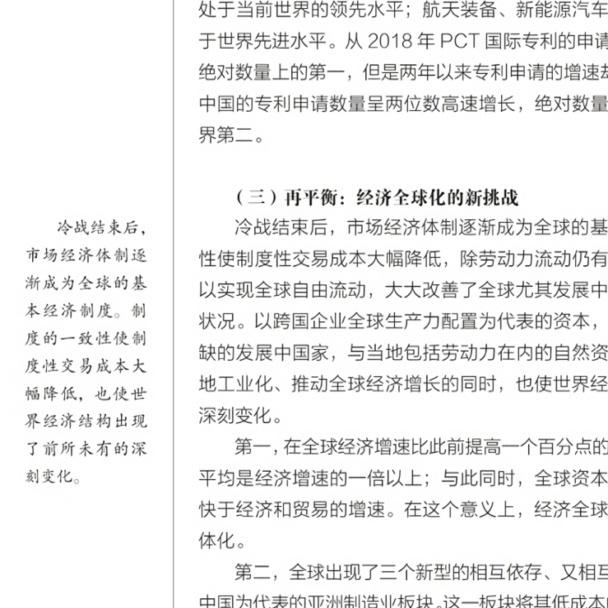}
    \caption{gt}
  \end{subfigure}
  \hfill
  \begin{subfigure}[b]{0.11\textwidth}
    \includegraphics[width=\textwidth]{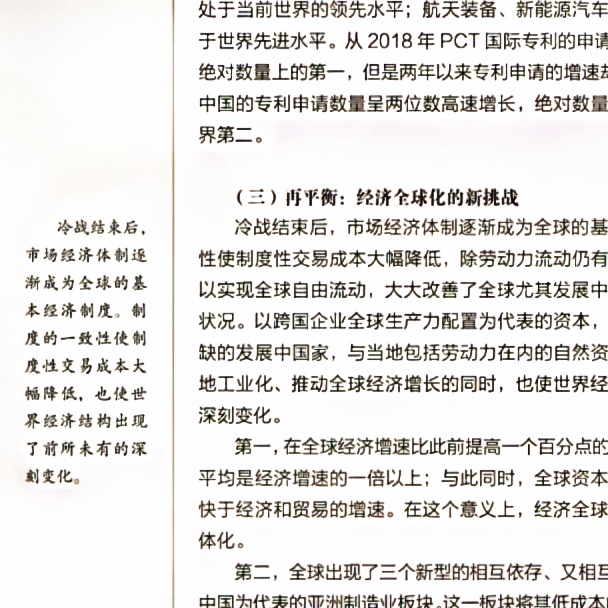}
    \caption{GCDRNet}
  \end{subfigure}
  \hfill
  \begin{subfigure}[b]{0.11\textwidth}
    \includegraphics[width=\textwidth]{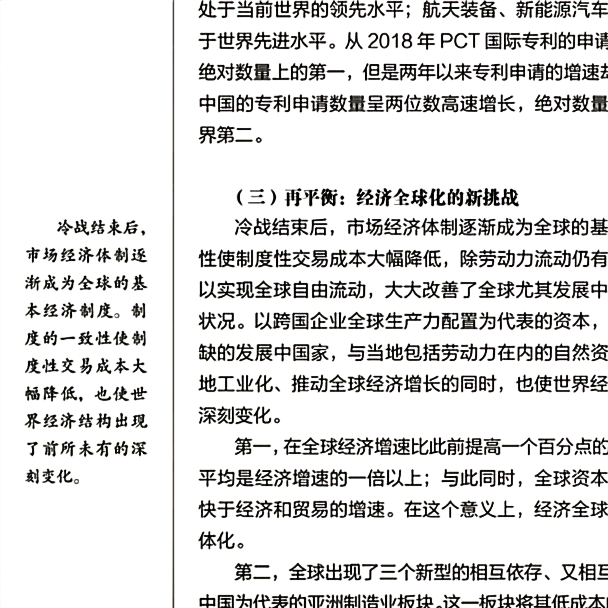}
    \caption{Our}
  \end{subfigure}
  
  \caption{\textbf{Cross-domain Visualization Comparison.} Qualitative comparison with GCDRNet demonstrates our method's superior performance in cross-domain scenarios. The visual evidence substantiates our approach's exceptional generalization capability, particularly in handling domain shift challenges.}
  \label{fig:other_dataset}
\end{figure}

\subsection{Objective Optimization}
Our comprehensive experimental analysis validates the critical importance of Structural Similarity (SSIM) Loss \cite{wang2004image} for maintaining fine-grained visual details in document image processing. The method demonstrates consistent effectiveness across tasks: super-resolution reconstruction, shadow removal, and image dewarping. To further enhance local structural consistency, we integrate Total Variation(TV) regularization \cite{aly2005image} as a complementary constraint. As evidenced by the systematic ablation study presented in Tab. \ref{tab:obj_func}, the progressive integration of SSIM loss and TV regularization yields measurable improvements in both SSIM and PSNR metrics.

\begin{table}[htbp]
\centering
\caption{Ablation study on objective functions for DIE.}\label{tab:obj_func}
\begin{tabular}{ccccc}
\toprule
  SSIM Loss & TV Loss & SSIM/PSNR \\
\midrule
            &         &    0.9077/22.78 \\
\checkmark  &         &    0.9193/23.56 \\
\checkmark  & \checkmark   &  0.9197/23.62 \\
\bottomrule
\end{tabular}
\end{table}

\subsection{Architectural Exploration of DB-LRNet}
We conducted systematic investigations into neural architecture design for DIE task. While RRDBNet \cite{wang2021real}, employing a columnar architecture, achieves state-of-the-art performance in super-resolution natural image, our frequency domain analysis (Fig.~\ref{fig:frequency_analysis}) reveals fundamental limitations in its direct application to document images enhancement. The document images exhibit richer high-frequency components compared to natural images, necessitating specialized architectural considerations. 

Our analysis identifies two critical design requirements: (1) enhanced feature reuse mechanisms for high-frequency detail preservation, and (2) sufficient receptive fields for robust distinguishing foreground-background. To address these requirements, we introduce a modified nested UNet \cite{zhou2018unet++} with dense block \cite{huang2017densely}, establishing a dense-in-dense structure. This design is motivated by two key insights: 
First, the dense connections facilitate improved gradient flow and feature propagation, reducing high-frequency information loss. Second, the multi-scale encoder-decoder structure enables contextual feature aggregation, which is crucial for handling complex document backgrounds.

Despite these improvements, conventional end-to-end generation approaches suffer from two inherent limitations: (1) compromised generalization performance on out-of-distribution samples, characterized by foreground degradation and regional artifacts; (2) unstable parameter convergence during gradient-based optimization. To address these challenges, we propose a dual-branch parametric transformation framework that replaces direct pixel prediction with smoothness-constrained coefficient learning. As detailed in Sec. \ref{sec:DB-LRNet}, DB-LRNet integrates an image smoothing branch and a linear transformation branch, generating enhanced outputs through learned parameters($\alpha$, $\beta$) and smooth image.

As quantified in Tab.~\ref{tab:arch_res}, our nested architecture achieves comparable performance to RRDBNet while significantly improving computational efficiency. The proposed NestUNet-Denseblock variant reduces parameter count by 51\% and computational complexity by 79\% compared to RRDBNet, establishing a new efficiency-performance frontier. Moreover, the proposed DB-LRNet achieves significant performance gains with minimal computational overhead, elevating SSIM from 0.9193 to 0.9385 while maintaining comparable efficiency (Tab.~\ref{tab:arch_res}). At the same time, this architecture demonstrates superior scalability for high-resolution document processing without performance degradation, as detailed in Sec. \ref{sec: efficient_inference}. 

\begin{table}[h]
\centering
\caption{\textbf{Comparative Analysis of Network Architectures on Image Quality and Computational Efficiency.} The RRDBNet$^{11}$ architecture is configured with 11 residual blocks, while the proposed NestUNet variant employs two distinct building blocks: VGG-style \cite{simonyan2014very} and Dense Block \cite{huang2017densely} modules. Notably, batch normalization layers are intentionally omitted from all architectures.}
\label{tab:arch_res}
\setlength{\tabcolsep}{3pt}
\begin{tabular}{cccc}
\toprule
Architecture  &  SSIM/PSNR     & Params(M) $\downarrow$ & GFLOPs $\downarrow$ \\
\midrule
 RRDBNet$^{11}$      & 0.9210/22.02           & 8.09                    & 38.25 \\
 NestUNet-VGG   & 0.9145/23.13           & 9.21                    & \textbf{3.71} \\
 NestUNet-Dense & 0.9193/23.56           & \textbf{3.99}           & 7.93 \\
 DB-LRNet            & \textbf{0.9385/23.99}  & 4.05                    & 8.12 \\
\bottomrule
\end{tabular}
\end{table}

\begin{table}[htbp]
\centering
\caption{\textbf{Comparative Analysis of Architectural Configurations.} Global module: GPPNet; Local module: DB-LRNet} 
\label{tab:two_stage}   
\begin{subtable}[t]{0.48\textwidth}
\centering
\begin{tabular}{cc}
\toprule
  Fusion Strategy & SSIM/PSNR \\
\midrule
Cascading      &  0.5781/12.36 \\
Additive       &  0.8717/18.55 \\
Concatenation  &  0.9111/21.92 \\
\bottomrule
\end{tabular}
\subcaption{Operation fusion strategies}
\label{tab:two_stage_suba}
\end{subtable}
\hfill 
\begin{subtable}[t]{0.48\textwidth}
\centering
\begin{tabular}{cc}
\toprule
Integration Strategy &  SSIM/PSNR \\
\midrule
Local  + Global  &  0.9231/23.44 \\
Global + Local   & 0.9480/24.10   \\
Global  &  0.8823/20.12    \\
\bottomrule
\end{tabular}
\subcaption{Stage integration strategies}
\label{tab:two_stage_subb}
\end{subtable}

\end{table}

\begin{table*}[!t]

\centering
\begin{subtable}[b]{0.45\textwidth}

\centering
\caption{Performance comparison on DocUNet dataset.}
\setlength{\tabcolsep}{3pt}
\begin{tabular}{lccc}
\toprule
\textbf{Venue} & \textbf{Methods}             & \textbf{SSIM} & \textbf{PSNR} \\
\midrule
TOG'19         & DocProj \cite{li2019document} & 0.7098         & 14.71 \\
BMVC'20        & Das et al. \cite{das2020intrinsic} & 0.7276    & 16.42 \\
MM'21          & DocTr \cite{feng2021doctr} & 0.7067            & 15.78 \\
MM'22          & UDoc-GAN \cite{wang2022udoc} & 0.6833          & 14.29 \\
TAI'23         & GCDRNet \cite{zhang2023appearance} & 0.7658    & 17.09 \\
CVPR'24        & DocRes \cite{Zhang_2024_CVPR} & 0.7598         & \textbf{17.60} \\
               & GL-PGENet(ours)          &   \textbf{0.7721}  & 16.89 \\
\bottomrule
\end{tabular}
\label{tab:sota_sub1}
\end{subtable}
\hfill
\begin{subtable}[b]{0.45\textwidth}

\centering
\caption{Performance comparison on RealDAE dataset.}
\setlength{\tabcolsep}{3pt}
\begin{tabular}{lccc}
\toprule
\textbf{Venue} & \textbf{Methods} & \textbf{SSIM}           & \textbf{PSNR} \\
\midrule
TOG'19         & DocProj \cite{li2019document} & 0.8684      & 19.35 \\
BMVC'20        & Das et al. \cite{das2020intrinsic} & 0.8633 & 19.87 \\
MM'21          & DocTr \cite{feng2021doctr} & 0.7925         & 18.62 \\
MM'22          & UDoc-GAN \cite{wang2022udoc} & 0.7558         & 16.43 \\
TAI'23         & GCDRNet \cite{zhang2023appearance} & 0.9423 & 24.42 \\
CVPR'24        & DocRes \cite{Zhang_2024_CVPR} & 0.9219      & \textbf{24.65} \\
               & GL-PGENet(ours)         &   \textbf{0.9480}& 24.10 \\
\bottomrule
\end{tabular}
\label{tab:sota_sub2}
\end{subtable}
\caption{\textbf{Quantitative comparison with state-of-the-art methods.} \textbf{Bold values} indicate best performance.}
\label{tab:sota}
\end{table*}

\subsection{Two-stage Enhancement Paradigm}
Our architectural refinements demonstrate substantial improvements in enhancement quality and cross-domain generalization. Nevertheless, achieving comprehensive consistency and robust generalization across heterogeneous document images remains challenging.

Inspired by GCDRNet \cite{zhang2023appearance}, we adopt a two-stage enhancement framework that systematically addresses document image enhancement. Unlike their UNet-based image transformation approach, our architecture implements a parametric regression network to optimize global enhancement through learned brightness, contrast, and saturation parameters. This global correction from GPPNet is subsequently refined by our proposed DB-LRNet in Sec. \ref{sec:DB-LRNet}) following a coarse-to-fine optimization strategy. Two critical implementation issues merit discussion: 1) feature integration of three parametric operations in GPPNet, and 2) optimal combination of global and local enhancement stages.

Tab. \ref{tab:two_stage} presents systematic comparisons of different architectural configurations. For GPPNet's operation fusion (Tab. \ref{tab:two_stage_suba}), concatenation achieves superior performance compared to cascading and additive fusion. Regarding stage combination (Tab. \ref{tab:two_stage_subb}), the Global + Local cascade paradigm yields optimal results, outperforming both Local + Global cascading and standalone global processing. These systematic evaluations validate our design choices for operation fusion and multi-stage coordination. The final GL-PGENet architecture, illustrated in Fig. \ref{fig:framework}, integrates these optimal configurations.

\subsection{Comparison with State-of-the-Art Methods}
We perform comprehensive evaluations of GL-PGENet against current state-of-the-art approaches on two benchmark datasets: RealDAE \cite{zhang2023appearance} and DocUNet \cite{ma2018docunet}. The RealDAE dataset, containing diverse real-world degradations, serves as a rigorous benchmark for evaluating robustness under complex practical conditions. For DocUNet evaluation, we adopt the standardized protocol from GCDRNet \cite{zhang2023appearance}, using geometrically aligned document images from the paper \cite{zhang2023docaligner} as degraded inputs to ensure pixel-wise correspondence with ground-truth images.

Our proposed GL-PGENet achieves state-of-the-art performance on both SSIM and PSNR metrics shown in Tab. \ref{tab:sota}, demonstrating consistent superiority in structural preservation. We argue that SSIM better correlates with human perceptual quality in document restoration tasks due to its sensitivity to structural distortions. On DocUNet dataset, GL-PGENet obtains an SSIM of 0.7721, outperforming DocRes \cite{Zhang_2024_CVPR} (0.7598) and GCDRNet \cite{zhang2023appearance} (0.7658). The significant SSIM improvement confirms enhanced capability in maintaining document structure and document readability. Our method also establishes a new state-of-the-art SSIM of 0.9480 while maintaining competitive PSNR performance on RealDAE dataset. 

We argue that the SSIM provides a more reliable indicator of human perceptual quality in document restoration tasks compared to pixel-wise metrics, owing to its enhanced sensitivity to structural distortions that critically affect document readability. Notably, our experimental results demonstrate superior performance in terms of SSIM compared to existing methods, thereby validating the effectiveness of our approach. Furthermore, qualitative comparisons provide visual evidence supporting this claim. As illustrated in Fig. \ref{fig:6x7grid}, our qualitative comparisons between GL-PGENet and state-of-the-art methods demonstrate its superior performance in enhancing local consistency while preserving fine-grained features. These comprehensive experimental results validate the effectiveness of our approach in addressing the challenging task of DIE.

\subsection{Efficient High-resolution Image Inference}
\label{sec: efficient_inference}
Our architecture demonstrates superior computational efficiency for high-resolution image processing through two complementary mechanisms. As depicted in Fig. \ref{fig:framework}, GPPNet employs fixed-resolution processing throughout its inference pipeline, ensuring computational stability regardless of input size. DB-LRNet employs a resolution reduction and coefficient map upsampling strategy for parameter prediction, achieving nearly $3 \times$ acceleration for high-resolution images (Fig. \ref{fig:framework_lrnet}). The lightweight smoothing branch maintains original resolution with minimal computational overhead.

Fig. \ref{fig:efficient_infer} provides visual evidence of our method's efficacy, where the baseline approach (top row) and our efficient implementation (bottom row) exhibit negligible performance degradation. Quantitative evaluations on the test set confirm this observation, showing only a 1.06\% relative decrease in SSIM (0.9480 vs. 0.9379) while maintaining significant superiority over DocRes \cite{Zhang_2024_CVPR}. This marginal performance trade-off enables substantial computational savings, making our approach particularly suitable for practical high-resolution image processing applications.

\subsection{Cross-domain Generalization}
The proposed method demonstrates strong cross-domain generalization capabilities, attributable to innovations in pre-training strategies and architectural design. As illustrated in Fig. \ref{fig:other_dataset}, comparative cross-domain evaluations between GL-PGENet and GCDRNet reveal significant performance advantages of our approach. The benchmark dataset encompasses multiple degradation types including shadows, blurs, and moire degradation, with a training set of 2,684 images and validation set of 123 images. For rigorous generalization assessment, we exclusively utilize the test set without any fine-tuning or additional training. The experimental results indicate that GL-PGENet exhibits significantly enhanced restoration performance, achieving an SSIM score of 0.8418 and a PSNR value of 18.47, which represents a notable improvement over GCDRNet's corresponding metrics of 0.8266 (SSIM) and 18.10 (PSNR). This marked performance gap substantiates the enhanced generalization capacity of our method across diverse degradation scenarios, suggesting better preservation of structural and photometric consistency in cross-domain dataset.

\section{Conclusion}
This paper introduces GL-PGENet, a novel framework designed to address the critical challenge of multi-degradation enhancement in color document images. Our approach makes four key contributions to Document AI research: 1) A hierarchical coarse-to-fine enhancement architecture that efficiently balances global consistency with local detail preservation; 2) A lightweight Global Perception Parameter Network that replaces computationally intensive pixel-wise estimation with efficient parametric regression for brightness, contrast, and saturation adjustments; 3) A Dual-Branch Local-Refine Network employing parameter generation mechanisms rather than direct pixel prediction, significantly improving generalization capabilities while maintaining local consistency; 4) A modified NestUNet architecture with integrated dense blocks specifically optimized for document image characteristics, effectively preserving high-frequency textual details crucial for downstream OCR tasks. 

Extensive experiments validate the effectiveness of our approach, achieving state-of-the-art SSIM scores of 0.7721 on DocUNet and 0.9480 on RealDAE, confirming superior structural preservation particularly critical for document readability. The proposed framework also demonstrates exceptional computational efficiency, reducing inference time by approximately 75\% for high-resolution documents without significant quality degradation, addressing a key limitation of existing transformer-based approaches. Furthermore, our model exhibits remarkable cross-domain adaptability, maintaining high performance (0.8418 SSIM) on unseen datasets without fine-tuning.

Looking forward, future research could extend this parametric generation paradigm to address additional document-specific degradations such as moire patterns and ink bleeding, while exploring self-supervised and weakly-supervised learning strategies to further reduce dependency on paired training data. By effectively balancing enhancement quality with computational efficiency, GL-PGENet provides a practical solution for real-world document digitization systems, serving as a robust foundation for downstream Document AI applications including OCR, layout analysis, and information extraction.

\bibliographystyle{aaai} 
\bibliography{references} 
\end{document}